\pdfoutput=1
\documentclass[11pt]{article}

\usepackage[preprint]{acl}

\usepackage{times}
\usepackage{latexsym}

\usepackage[T1]{fontenc}
\usepackage[utf8]{inputenc}

\usepackage{microtype}

\usepackage{inconsolata}

\usepackage{graphicx}
\usepackage{hyperref}       % hyperlinks
\usepackage{url}            % simple URL typesetting
\usepackage{booktabs}       % professional-quality tables
\usepackage{amsfonts}       % blackboard math symbols
\usepackage{nicefrac}       % compact symbols for 1/2, etc.
\usepackage{microtype}      % microtypography
\usepackage{graphicx}
\usepackage{caption}
\usepackage{booktabs}
\usepackage{amsmath}
\usepackage{amssymb}
\usepackage{pifont}
\usepackage{subcaption}
\usepackage{stfloats}
\usepackage{wrapfig}
\usepackage{multirow}
\usepackage{hyperref}

\title{Towards Better Chain-of-Thought: A Reflection on \\ Effectiveness and Faithfulness\\ }

\author{Jiachun Li\textsuperscript{1,2}, Pengfei Cao\textsuperscript{1,2}\footnotemark[1], Yubo Chen\textsuperscript{1,2} Jiexin Xu\textsuperscript{3}, Huaijun Li\textsuperscript{3}, \\ \textbf{Xiaojian Jiang}\textsuperscript{3}, \textbf{Kang Liu\textsuperscript{1,2}, Jun Zhao\textsuperscript{1,2}\footnotemark[1]} \\ \textsuperscript{1}School of Artificial Intelligence, University of Chinese Academy of Sciences \\ \textsuperscript{2}The Key Laboratory of Cognition and Decision Intelligence for Complex Systems, \\ Institute of Automation, Chinese Academy of Sciences  \\ \textsuperscript{3}China Merchants Bank\\
\footnotesize{\texttt{\{jiachun.li, pengfei.cao, yubo.chen, kliu, jzhao\}@nlpr.ia.ac.cn }}}

\begin{document}
\maketitle
\renewcommand{\thefootnote}{\fnsymbol{footnote}}
\footnotetext[1]{Corresponding authors.}
\renewcommand{\thefootnote}{\arabic{footnote}}
\begin{abstract}
Chain-of-thought (CoT) prompting demonstrates varying performance under different reasoning tasks.
Previous work attempts to evaluate it but falls short in providing an in-depth analysis of patterns that influence the CoT. In this paper, we study the CoT performance from the perspective of effectiveness and faithfulness. For the former, we identify key factors that influence CoT effectiveness on performance improvement, including problem difficulty, information gain, and information flow. For the latter, we interpret the unfaithful CoT issue by conducting a joint analysis of the information interaction among the question, CoT, and answer. The result demonstrates that, when the LLM predicts answers, it can recall correct information missing in the CoT from the question, leading to the problem. Finally, we propose a novel algorithm to mitigate this issue, in which we recall extra information from the question to enhance the CoT generation and evaluate CoTs based on their information gain. Extensive experiments demonstrate that our approach enhances both the faithfulness and effectiveness of CoT.
\end{abstract}

\section{Introduction}
Recently, with chain-of-thought (CoT) techniques \citep{cot}, large language models (LLMs) are able to reason on complex tasks \citep{cot-sc, gpt4, rag_reward, linked}. By scaling the CoT process using reinforcement learning (RL), LLMs can even surpass human performance in competition-level mathematical problems \citep{o1, deepseek-r1}. However, despite the significant success of the CoT, some studies find that it demonstrates poor performance on certain tasks \citep{cot_or_not, llm_paradox, unfaithful_cot, measure_cot_faith}. In some cases, using CoT for the model's reasoning is unnecessary or even harmful \citep{cot_probe, toxic_cot}. 

These conflicting findings motivate the need for a systematic analysis of the CoT. To this end, a series of studies evaluating CoT's performance has commenced \citep{unfaithful_cot, llm_causal, cot_probe, measure_cot_faith}, which can be mainly divided into two lines: 
On the one hand, some works assess CoT based on its effectiveness. They measure the accuracy improvements brought by the CoT across different tasks and identify task types where CoT is effective \citep{cot_or_not, llm_paradox, cot_effect}.
On the other hand, some works evaluate the CoT based on its faithfulness \citep{faith_define, faith_test}.  They investigate the consistency between CoTs and final answers by analyzing the causal relevance linking them. \citep{measure_cot_faith, faith_sc, llm_causal}. Effectiveness is result-oriented, focusing on whether CoT can enhance the quality of reasoning outcomes; whereas faithfulness is process-oriented, concerned with whether the reasoning process of CoT genuinely influences the results. 

Though these works have made great progress, they lack an in-depth analysis of the patterns influencing CoT performance.
For the effectiveness evaluation works, they draw conclusions like CoT performs well in tasks involving mathematical symbols, but does not explore the underlying factors influencing these conclusions \citep{cot_or_not, llm_paradox}.
For the faithfulness evaluation works, they primarily design various methods to determine whether CoT is faithful, but lack an explanation for the issue of CoT unfaithfulness.
\citep{faith_cot, measure_cot_faith, llm_causal}.

\textbf{In this paper, we focus on analyzing key patterns that influence the CoT's performance from both effectiveness and faithfulness perspectives.}
From the effectiveness perspective, we identify three factors that contribute to CoT's final improvement, including problem difficulty, information gain, and information flow. 
We start by splitting questions into various difficulty levels and comparing the model's accuracy on them, from which we find that CoT is more effective on harder problems. 
Then, we calculate the information gain brought by CoT for questions across different tasks and demonstrate CoT with more additional information tends to be more effective.
Lastly, we consider the internal information flow during model reasoning. Through the experiment, we conclude that the more information interaction increases with the CoT process, the more effective the CoT becomes.
From the faithfulness perspective, we discover that there exist non-negligible unfaithful CoT issues in logical reasoning, where an incorrect CoT can still lead to the correct answer. We further interpret this issue by jointly analyzing the information interaction among question, CoT, and answer. Through it, we identify three patterns that lead to the CoT's unfaithfulness: (1) CoT loses key information from the question; (2) CoT transfers less information to the answer; (3) The model recalls correct information from the question when answering.

At last, we explore the relationship between the above two perspectives. A novel algorithm called \textbf{QU}estion
\textbf{I}nformation \textbf{R}ecall and \textbf{E}nhancement (QUIRE) is designed to mitigate the unfaithful CoT issue. In it, we first generate a raw answer to recall correct information from the question, then use this extra information to prompt the generation of a new CoT generation. Finally, we employ the CoT information gain as the weight to vote for the final answer. Through extensive experiments, we not only demonstrate that our method can mitigate unfaithful issues, but also show that CoT faithfulness is a key factor in influencing CoT effectiveness.

In summary, our key contributions are as follows:
(1) We identify key factors that influence CoT's effectiveness on different reasoning tasks, including problem difficulty, information gain, and information flow. 
(2) We interpret the unfaithful CoT issue by jointly analyzing the information interaction among question, CoT, and answer. Based on experimental results, we demonstrate that the reason is that LLMs retrieve correct information (lost in the CoT) directly from the question when predicting answers.
(3) As an application of our findings, we design a new method called QUIRE, which effectively improves the CoT's performance from the effectiveness (up to \textbf{2.4\%} improvement) and faithfulness (up to \textbf{5.6\%} improvement). This indicates that enhancing CoT faithfulness can lead to an improvement in CoT effectiveness. Our code is available at: 
\href{https://github.com/BugMakerzzz/better_cot}{https://github.com/BugMakerzzz/better\_cot}.

\section{Related Works}
\subsection{Chain-of-Thought Effectiveness}
Since the emergence of CoT, a series of CoT-like approaches have further improved the model's reasoning accuracy through various prompt designs \citep{cot-sc, sr, l2m}. Recently, the emergence of reasoning models such as DeepSeek-R1 \citep{deepseek-r1} and o1 \citep{o1} has once again proven that CoT is highly effective in solving complex reasoning tasks such as mathematics and coding \citep{rstar, scaling-test, scale_survey, verify-step}. However, another series of works shows that the effectiveness of CoT has limitations \citep{cot_probe, llm_paradox, mirage}. They demonstrate that CoT brings only limited improvements in knowledge and commonsense reasoning tasks \citep{cot_or_not}, and may even harm the model's original performance \citep{toxic_cot}. Building on these studies, our work further investigates the key factors that control CoT's effectiveness across different tasks.

\subsection{Chain-of-Thought Faithfulness}
In model interpretability, faithfulness, defined as ``accurately representing the reasoning process behind the model’s decision'', is important for evaluating the performance of natural language explanation \citep{interpret_1,interpret_2,faith_define}. With the emergence of CoT-like work, there has been increasing focus on measuring this characteristic within CoTs \citep{unfaithful_cot,measure_cot_faith,faith_cot}. Some studies introduce counterfactual perturbations to questions and measure the change of answers \citep{faith_test,unfaithful_cot}. Some other works use causal median analysis on CoTs and answers, calculating the treatment effect to represent the faithfulness \citep{llm_causal,improve_faith}. However, these works lack a comprehensive explanation and mitigation of unfaithful CoT, and this paper addresses this gap.

\section{What Makes CoT Effective} \label{sec:3}
In this section, we investigate what factors make the CoT effective in certain reasoning tasks. Specifically, we start with evaluating the final accuracy improvement of CoT on different tasks ($\S$\ref{sec:3.1}). Then we study the impact of three different factors on the final performance of CoT, including problem difficulty ($\S$\ref{sec:3.2}), CoT information gain ($\S$\ref{sec:3.3}), and the information flow between CoT and answer ($\S$\ref{sec:3.4}).

\subsection{Overall Performance}\label{sec:3.1}
\paragraph{Experimental Setup} We choose 9 representative datasets from various reasoning types for evaluation.
Specifically, for mathematical reasoning, we choose GSMIC \citep{gsmic}, GSM8K (GSM) \citep{GSMK} and AQuA \citep{aqua}.
For logical reasoning, we choose ProofWriter (PW) \citep{proofwriter}, FOLIO \citep{folio} and ProntoQA (PQA) \citep{prontoqa}.
For commonsense reasoning, we choose WinoGrande (WINO) \citep{wino}, SocialIQA (SIQA) \citep{siqa} and ECQA \citep{ecqa}.
For models, due to the difficulty of deeply analyzing the internals of black-box models, we focus on analyzing white-box models and select four advanced white-box LLMs for the experiment, including Mistral-7B \citep{mistral}, Gemma2-9B \citep{gemma2}, Llama3.1-8B \citep{llama3}, and Qwen2.5-14B \citep{qwen2.5}. 
For metrics, we define the effectiveness of CoT as the difference in accuracy when answering questions with and without CoT.

\begin{figure}[tbp]

    \centering
    \includegraphics[width=\linewidth]{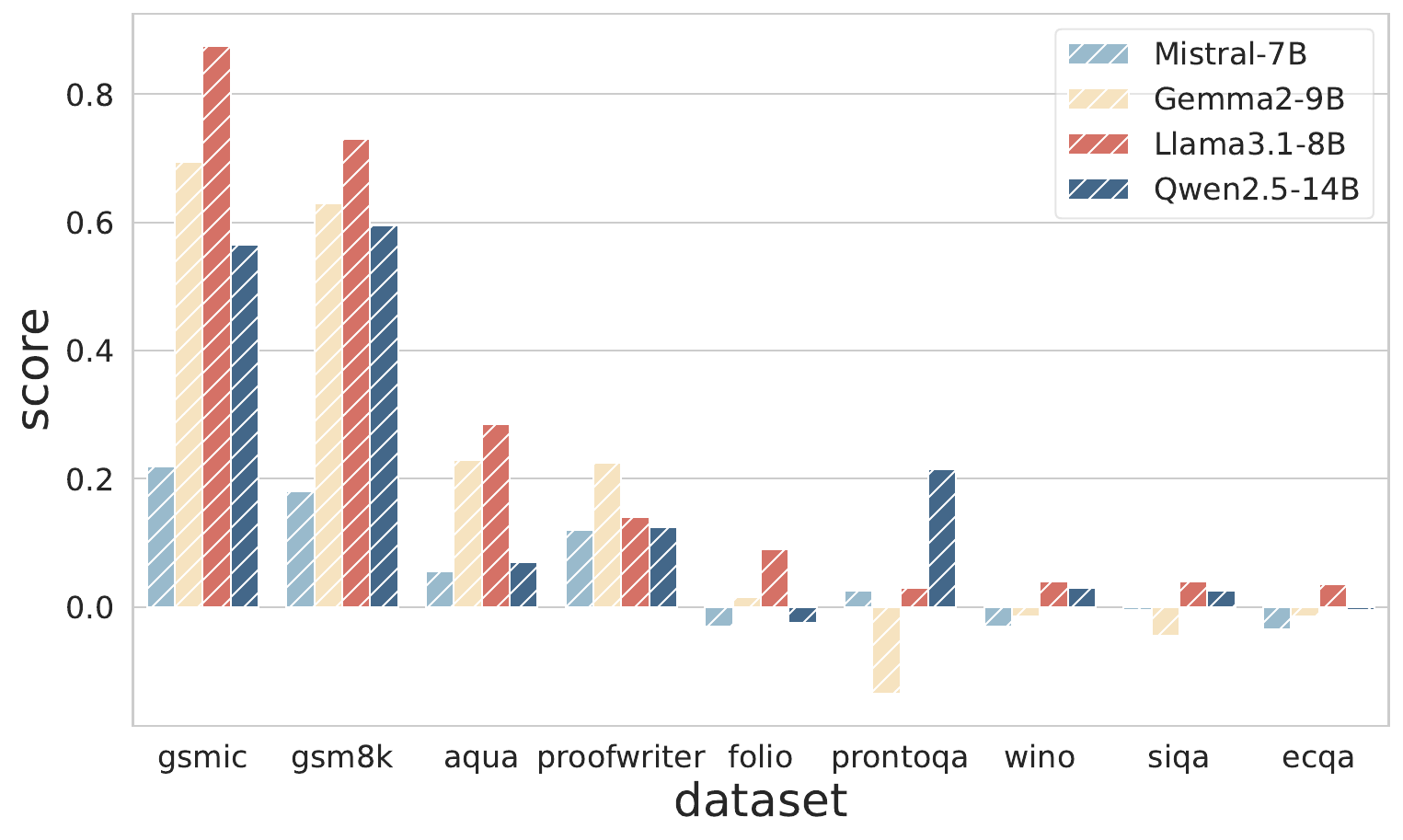}
    \caption{CoT improvement across different models and datasets, `score' indicates the accuracy difference.}
    \label{fig:perform_acc}

\end{figure}

\paragraph{Main Results}
The main results of the evaluation experiment are illustrated in Figure \ref{fig:perform_acc}, from which we can get that: \textbf{Among different reasoning tasks, CoT is most effective in mathematical reasoning, while least effective in commonsense reasoning tasks.} This conclusion forms the basis for the subsequent analysis in this section.

\subsection{Problem Difficulty} \label{sec:3.2}
Why is CoT more effective on certain task types? Reflecting on humans' reasoning process, the more difficult the problem, the more thinking time is required. Hence, we aim to explore whether this pattern can also be observed in LLMs: Is CoT more effective for harder problems?
\paragraph{Problem Difficulty Estimation} Following former works, we classify the difficulty of questions based on the model's accuracy in answering them \citep{verify-step, reward_progress}. 
Specifically, for each question, we sample 10 answers without CoT prompting and bin the average pass@1 rate across all models into five quantiles, each corresponding to increasing difficulty levels. For example, if the pass@1 rate is less than 0.1, the question is classified as the hardest level 5. Conversely, if the pass@1 rate is more than 0.8, the question is classified as the easiest level 1. 

\begin{figure}[tbp]

    \centering
 %    \subfigbottomskip=2pt %两行子图之间的行间距
	% \subfigcapskip=-5pt %设置子图与子标题之间的距离
    \begin{subfigure}[t]{.49\linewidth}
        \centering
	\includegraphics[width=\linewidth]{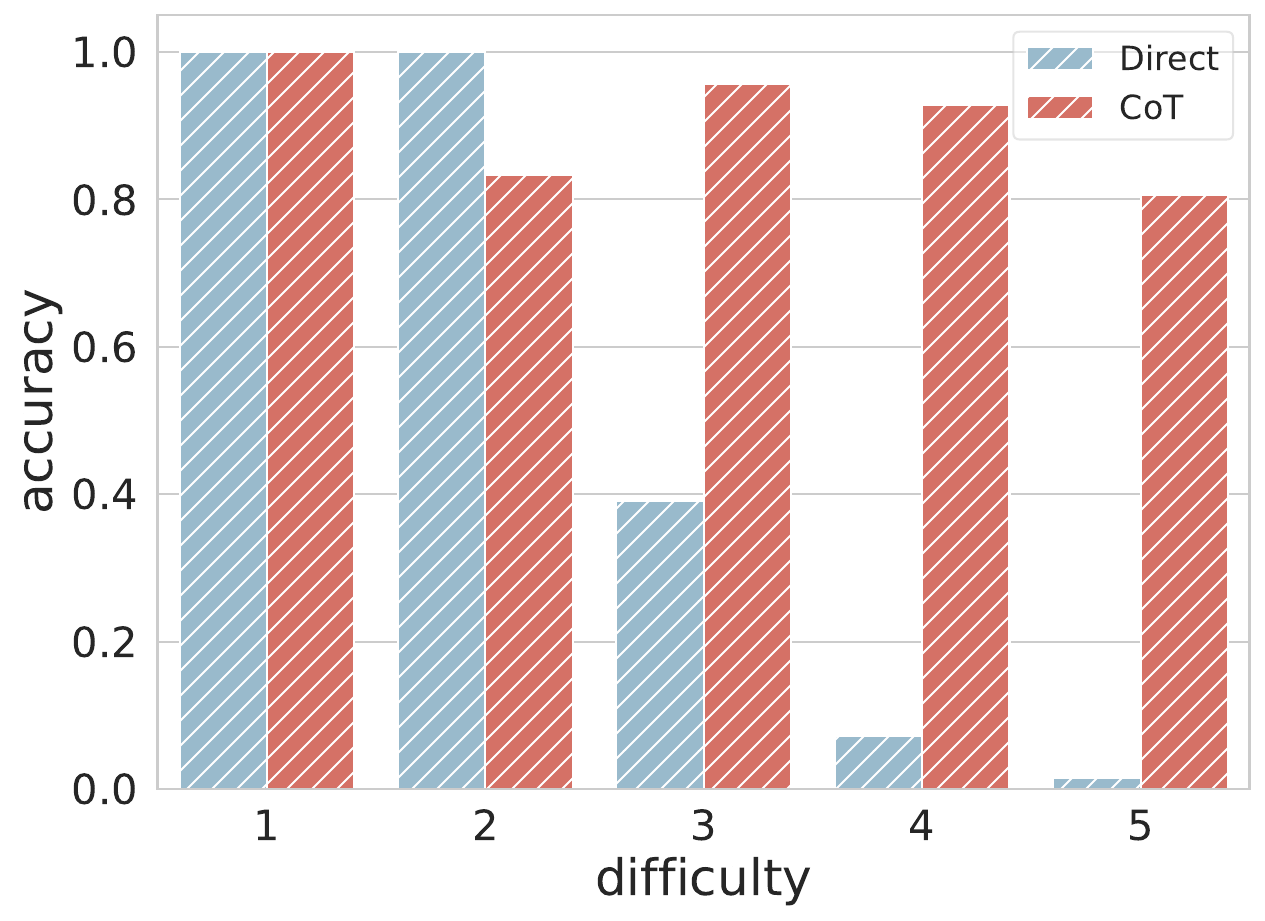}
        \caption{GSM8k}\label{fig:gsm8k_dif}
    \end{subfigure}
       \begin{subfigure}[t]{.49\linewidth}
        \centering
	\includegraphics[width=\linewidth]{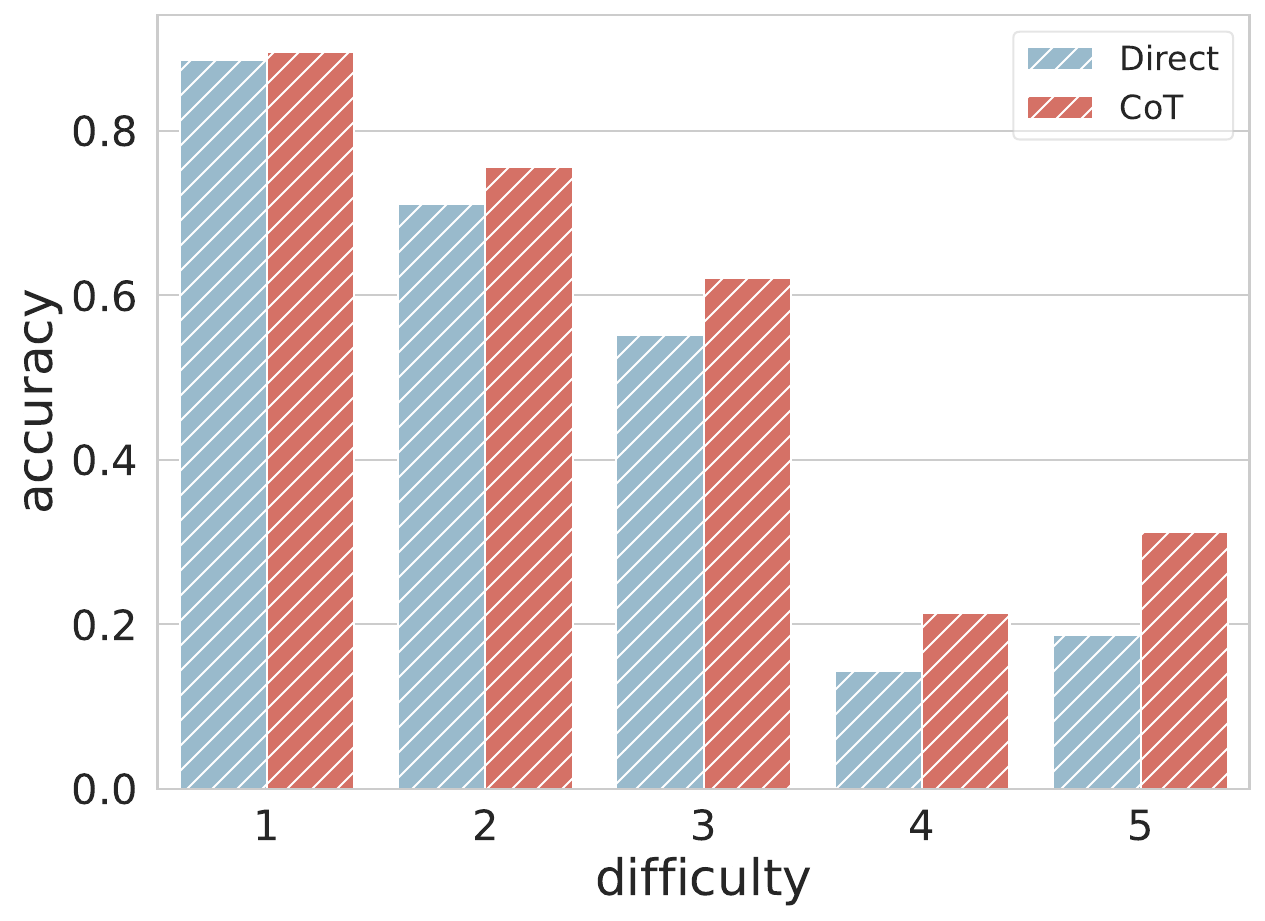}
        \caption{WinoGrande}\label{fig:dist_reason_llama}
    \end{subfigure}
    \\
    \caption{Performance on different problem difficulty levels with and without CoT prompting (Llama3.1-8B).}
    \label{fig:dif_acc}
% \end{wrapfigure}

\end{figure}

\paragraph{Performance across Difficulty Levels} After classifying the question, we compare the effectiveness of CoT across different difficulty levels and illustrate part of the results in Figure \ref{fig:dif_acc} (more results in Appendix \ref{append:diff}). We can conclude that: \hypertarget{Cl.1}{\textbf{(Cl.1) CoT is more effective on challenging questions compared to simple ones.}} For questions at low difficulty levels (e.g. level 1, level 2), CoT provides minimal accuracy improvement and even degrades performance. In contrast, CoT significantly increases reasoning accuracy across different tasks when the question is difficult (e.g. level 4, level 5).

\begin{figure}[tbp]

    \centering
    \includegraphics[width=0.9\linewidth]{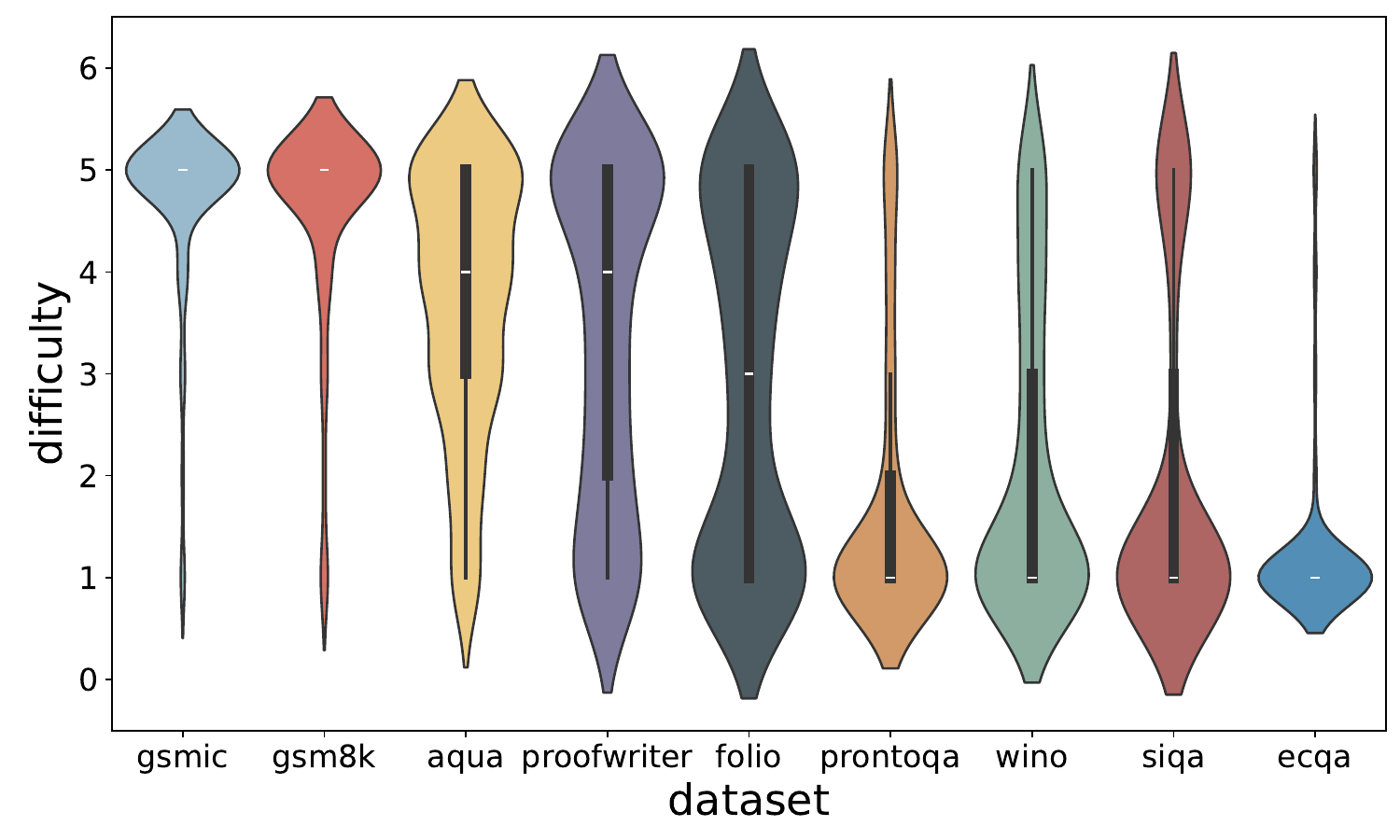}
    \caption{Difficulty distribution in different datasets.}
    \label{fig:dif_stat}
% \end{wrapfigure}

\end{figure}

\paragraph{Difficulty Distribution on Different Tasks} We further evaluate the difficulty distribution of different tasks to explain the varying effectiveness. Figure \ref{fig:dif_stat} shows the results on Llama3.1-8B. In mathematical reasoning, most problems are of higher difficulty, whereas in commonsense reasoning, most problems are of lower difficulty. Combining \hyperlink{Cl.1}{Cl.1}, we can infer that the CoT is more effective in mathematical reasoning since it has more difficult problems compared to other tasks. This provides an explanation for the effectiveness distribution shown in Figure \ref{fig:perform_acc} from the perspective of problem difficulty.

\subsection{Information Gain} \label{sec:3.3}
When we define the problem difficulty, we only consider the final result of LLM's reasoning. To conduct a more comprehensive analysis, we delve into the reasoning process and continue to identify key factors.
In practice, a harder question tends to require more extra information to answer. Thus, here we focus on the information gain of CoT in the reasoning process.

\paragraph{Information Gain Definition}
In information theory, Information Gain (IG) quantifies the reduction in uncertainty of the target variable $Y$ after adding a certain feature $X$:
\begin{equation}
 \footnotesize{
IG(Y,X)= H(Y) - H(Y|X)
}
\end{equation}
where $H(Y)$ represents the entropy of $Y$, and $H(Y|X)$ represents the conditional entropy of $Y$ given the feature $X$. Similarly, in the context of LLM reasoning, given a question $Q$ and a CoT $C$, we define the IG as follows:
\begin{equation}\label{eq:ig}
 \footnotesize{
\begin{aligned}
IG(C,Q) &= H(C) - H(C|Q) \\
 &= - \sum_{i=1}^{n} p(c_i|C_{i-1}) \log p(c_i|C_{i-1})  \\
 & + \sum_{i=1}^{n} p(c_i|C_{i-1};Q) \log p(c_i|C_{i-1};Q)
\end{aligned}
}
\end{equation}
Here, $p(\cdot)$ indicates the model's output probability, $C_{i-1}$ is the first $i-1$ tokens of CoT, and $n$ is the length of CoT.
IG represents the degree to which the uncertainty of CoT is reduced by the question. The larger the IG,  the more information CoT obtains from the question, hence the less additional information is provided by CoT itself.

\begin{figure}[tbp]

    \centering
    \includegraphics[width=0.9\linewidth]{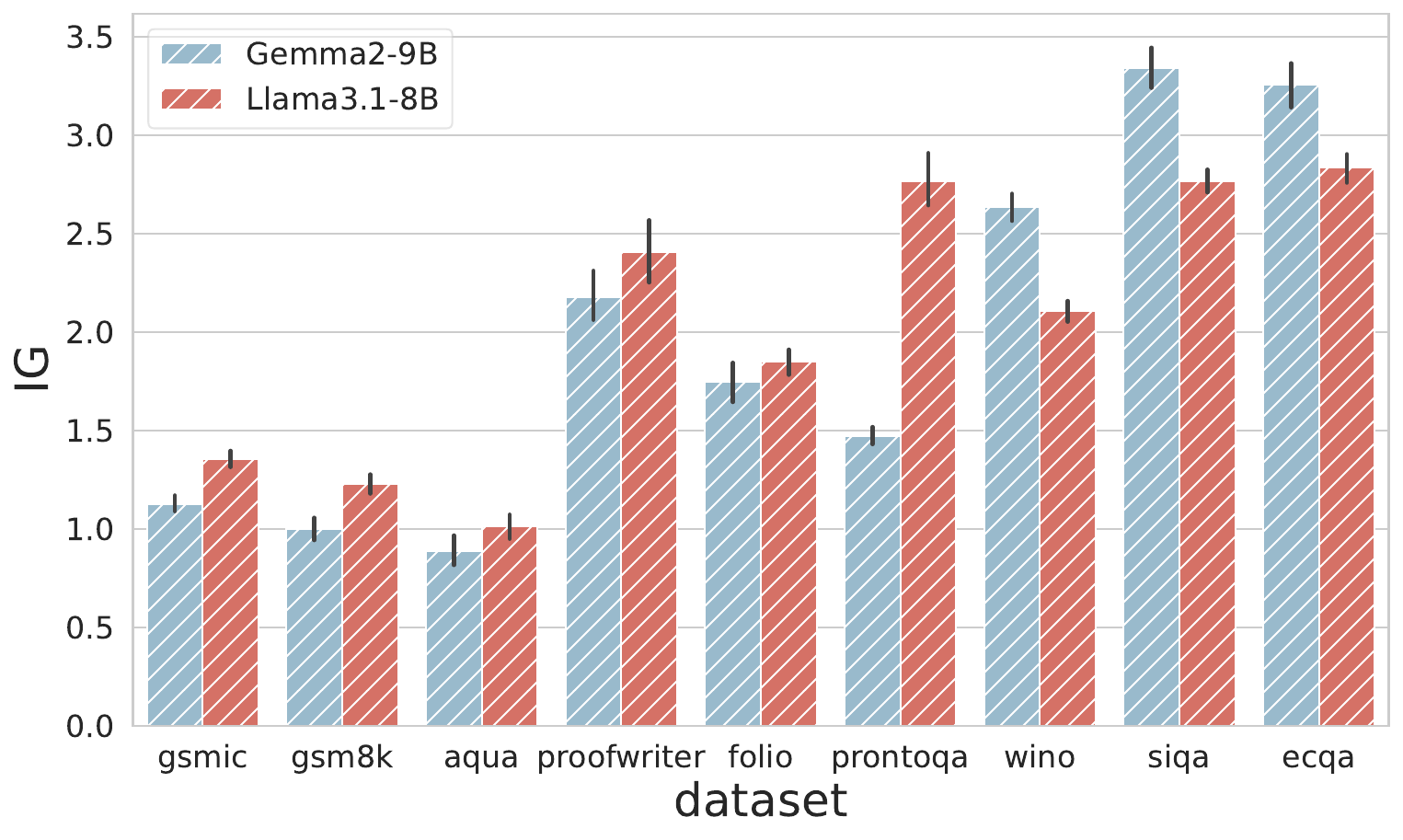}
    \caption{CoT information gain in different datasets.}
    \label{fig:ig}
% \end{wrapfigure}

\end{figure}

\paragraph{Experiment and Analysis} We conduct experiments across different datasets and demonstrate the results in Figure \ref{fig:ig}. Compared to Figure \ref{fig:perform_acc}, this figure shows an opposite trend: mathematical reasoning has the lowest IG, while commonsense tasks exhibit the highest IG. This indicates that: \hypertarget{Cl,2}{\textbf{(Cl.2) CoT is more effective when it provides additional information not present in the problem itself}} (e.g. gsmic, gsm8k, aqua). In contrast, when CoT is ineffective for performance improvement, it provides less extra information.

\subsection{Information Flow} \label{sec:3.4}
 In $\S$ \ref{sec:3.3}, we primarily demonstrate the importance of additional information in CoT. However, does the way in which models utilize this information also affect the CoT effectiveness? To answer this question, we study the information flow between CoT and answers in this experiment. 

\paragraph{Information Tracing Method} 
Following previous works \citep{gradient-tracing, gradient-xai, toxic_cot, cut_head}, we employ integrated gradient attribution (IGA) \citep{ig} as our measuring method to capture the information flow between CoT and answer. Specifically, we first compute importance $I_{n,m}$ of input token $x_{n}$ to output token $y_{m}$ as follows:
\begin{equation}\label{eq:information}
   \footnotesize{
   \begin{aligned}
I(x_n,y_m) &= E(x_n) \int_{\alpha=0}^{1} \frac{\partial f(\alpha y_m)}{\partial E(x_n)} d \alpha   \\
&\approx \frac{E(x_n)}{m} \sum_{k=1}^{m} \frac{\partial f(\frac{k}{m} y_m )}{\partial E(x_n)}
    \end{aligned}}
\end{equation}
where $f(\cdot)$ represents the model's output probability, $E(x_n)$ is the input word embedding of the token $x_n$ and $m$ is the number of approximation steps (we set it to 20). To reduce the interference from noise, we rescale the importance and get the attribution effect score between $x_{n}$ and $y_{m}$:
\begin{equation}\label{eq:attribution}
   \footnotesize{
AE(x_n,y_m) = \left\{\begin{array}{lr}
            \frac{I(x_n,y_m)}{\max_{n'=1}^N I(x_n',y_m)} & I(x_n,y_m) > 0 \\
            0 & otherwise 
    \end{array}
\right.}
\end{equation}
Here $N$ is the last index of the input. Finally, we can measure the information flow between each token $c$ of CoT and the answer $A$ using the average attribution effect (AAE): \begin{equation}\label{eq:aae}
   \footnotesize{
   \begin{aligned}
AAE(c,A) = \frac{1}{|A|} \sum_{a \in A} AE(c,a)
    \end{aligned}}
\end{equation}
Since CoT is usually long, averaging over each token of CoT would result in a significant loss of information. Hence, we choose to average over $A$ and analyze how the information flow changes throughout the CoT process using the AAE.

\begin{figure}[tbp]
     \centering
 %    \subfigbottomskip=2pt %两行子图之间的行间距
	% \subfigcapskip=-5pt %设置子图与子标题之间的距离
    \begin{subfigure}[t]{.49\linewidth}
        \centering
	\includegraphics[width=\linewidth]{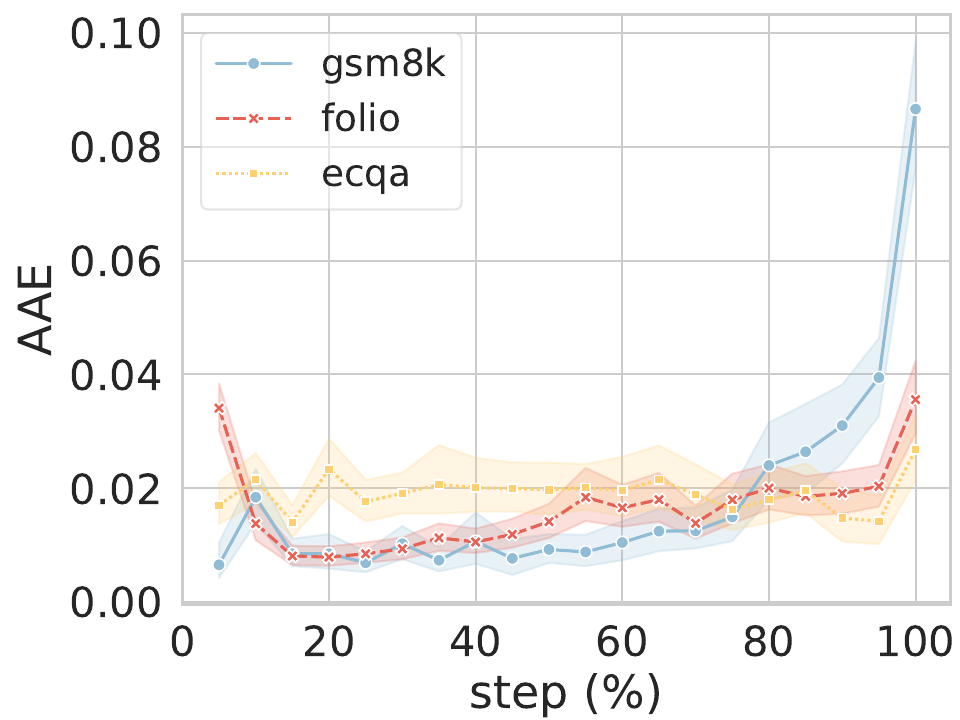}
        \caption{Gemma2-9B}\label{fig:gsm8k_dif}
    \end{subfigure}
       \begin{subfigure}[t]{.49\linewidth}
        \centering
	\includegraphics[width=\linewidth]{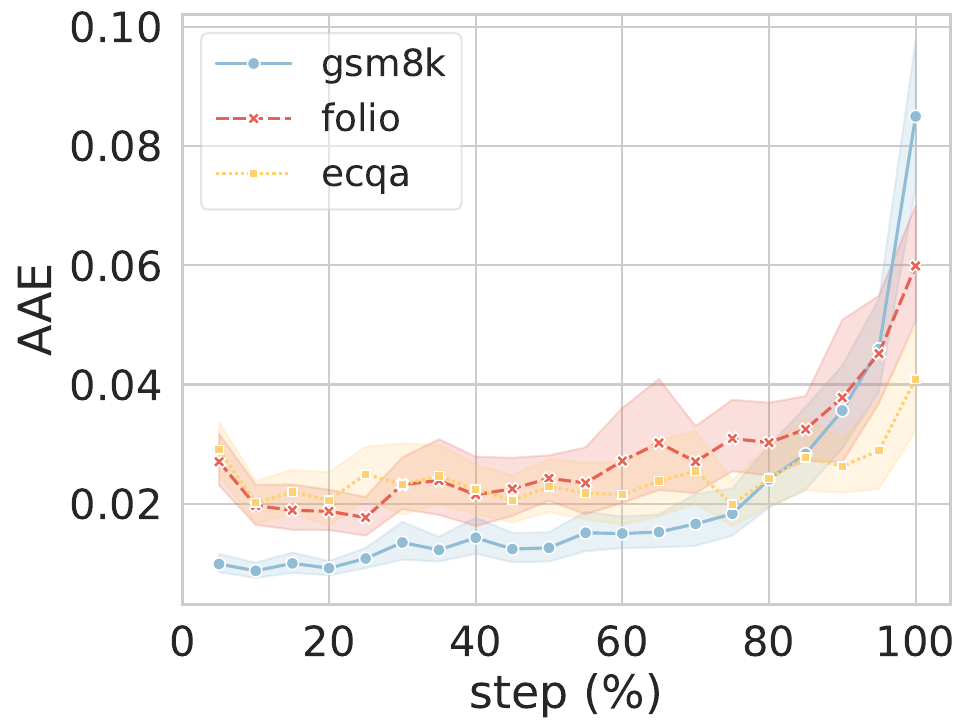}
        \caption{Llama3.1-8B}\label{fig:dist_reason_llama}
    \end{subfigure}
    \\
    \caption{Information flow between the CoT and answer. `Step' indicates sequential positions within the CoT, where 0 is the beginning and 100 is the end.}
    \label{fig:aae}

\end{figure}

\paragraph{Information Flow Comparison} 
We collect 200 CoT-answer pairs from three different datasets to calculate the AAE. Figure \ref{fig:aae} shows the main results, from which we can get that: \hypertarget{Cl.3}{}\textbf{(Cl.3) When information flow between CoT and the answer increases with the CoT process, the CoT tends to be effective.} As we can see from Figure \ref{fig:aae}, the curve of GSM8k exhibits the most significant upward trend, while ECQA remains the most stable, with the AAE showing little variation as the steps change. 
For tasks where CoT is highly effective (e.g. GSM8k), the influence of the CoT on the answer increases as the reasoning progresses. In contrast, for tasks that CoT is ineffective (e.g. ECQA), the influence of CoT on the answer does not significantly change as the reasoning progresses.
\begin{figure}[tbp]
    \centering
    \includegraphics[width=0.9\linewidth]{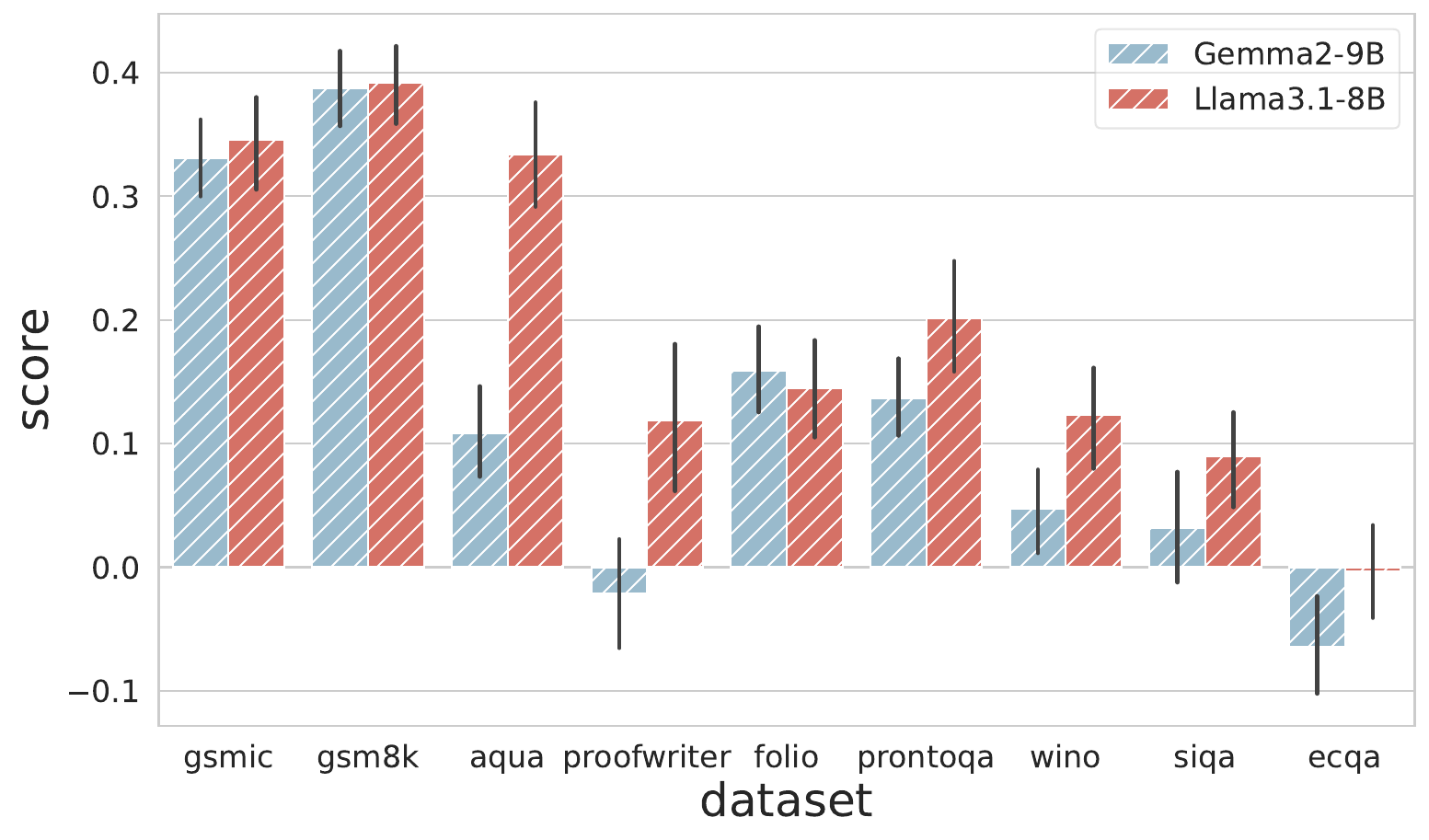}
    \caption{MIF score in different datasets.}
    \label{fig:mif}
\end{figure}

\paragraph{Monotonicity of Information Flow}
In the previous experiment, we identify the influence of AAE's increase by observing different curves. To quantitatively measure this increase, we define the monotonicity of information flow (MIF) as the Spearman correlation coefficient between the steps and the corresponding AAE values:
\begin{equation}\label{eq:information}
   \footnotesize{
   \begin{aligned}
MIF(C,A) &=  1 - \frac{6 \sum d_i^2}{n(n^2 - 1)} \\
&= 1 - \frac{6 \sum_{i=1}^{n} [n+1-i - R(AAE(c_i,A))]^2}{n(n^2 - 1)}
    \end{aligned}}
\end{equation}
where $n$ is the length of CoT and $R(\cdot)$ is the ranking of the value. In the implementation, we merge adjacent tokens and calculate their average AAE, thereby reducing noise interference. The experimental results on Gemma2-9B and Llama3.1-8B are presented in Figure \ref{fig:mif}, from which we can get that: 
\textbf{The higher the monotonicity of the information transfer between CoT and the answer, the more effective the CoT becomes.} This further demonstrates the validity of \hyperlink{Cl.3}{Cl.3}.

\section{What Makes CoT Unfaithful}\label{sec:4}
% In $\S$\ref{sec:3}, we evaluate effectiveness from the perspective of final accuracy, without considering the correctness of CoT itself. However, if the CoT itself is incorrect, then even if the answer is correct, it should still be considered ineffective.
% Following previous works \citep{faith_cot, measure_cot_faith, unfaithful_cot}, we name this type of CoT ineffectiveness, which is not covered in the previous analysis, as \textbf{``unfaithful CoT''.}
\begin{figure*}[tbp] 
    \centering
	\includegraphics[width=0.9\linewidth]{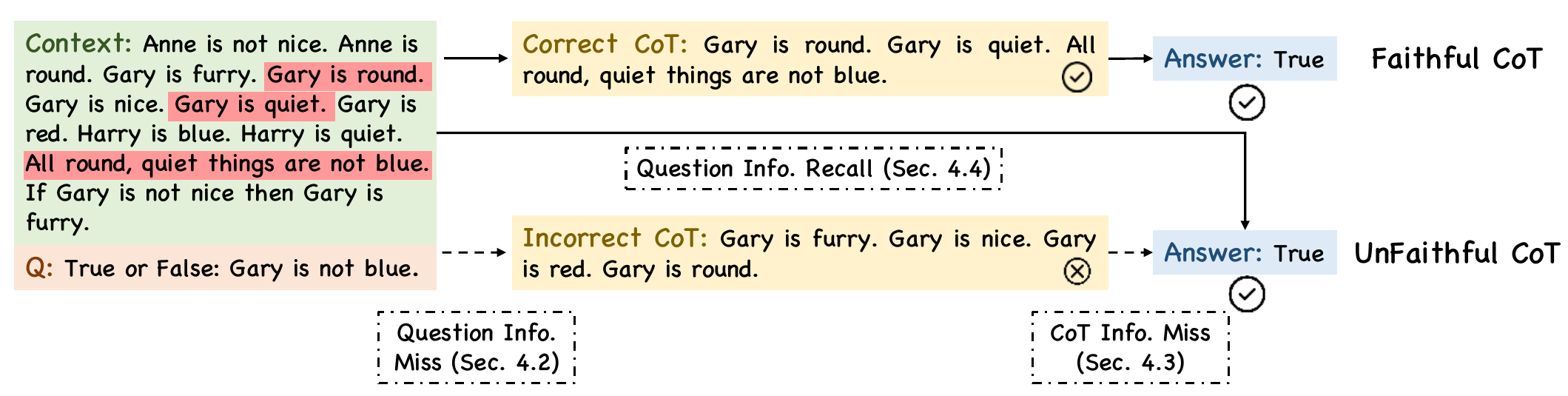}
    \caption{An interpretation of unfaithful CoT issues, where statements in red are correct information for reasoning. }
    \label{fig:bridge}
\end{figure*}

In this section, we aim to analyze the CoT from the faithfulness perspective.
Concretely, we first identify the unfaithfulness problem in different tasks ($\S$\ref{sec:4.1}). Next, we analyze the issue by examining the information interaction among the three key components of reasoning (as illustrated in Figure \ref{fig:bridge}), including question and CoT ($\S$\ref{sec:4.2}), CoT and answer ($\S$\ref{sec:4.3}), question and answer ($\S$\ref{sec:4.4}).

\begin{table}[t]
    \centering
     \scalebox{0.7}{
    \begin{tabular}{lcccccc} 
            \toprule
            C. $\rightarrow$ A.  & GSM & AQuA & PW & PQA & WINO & SIQA\\
            \midrule
             \ding{51} $\rightarrow$ \ding{51}& 41 & 25 & 14 & 27 & 34 & 40\\
             \ding{51} $\rightarrow$ \ding{55}&  0 & 0 & 0 & 0 & 1 & 0\\
             \ding{55} $\rightarrow$ \ding{51}& 1 & 1 & 7 & 17 & 1 & 0\\
             \ding{55} $\rightarrow$ \ding{55}& 8 & 24 & 29 & 6 & 14 & 10\\
            \bottomrule
    \end{tabular}}
        \caption{Inconsistency statistics between the CoT (C.) and the answer (A.) on Llama3.1-8B.}\label{tab:paradigm_faith}  
\end{table}
\subsection{CoT Faithfulness Evaluation} \label{sec:4.1}
Following previous works \citep{llm_causal, faith_cot}, we evaluate the faithfulness of CoT by measuring the consistency between the CoT and the answer. If an incorrect CoT induces a correct answer or a correct CoT induces a wrong answer, it is seen as an unfaithful CoT (see Figure \ref{fig:bridge} for example). We manually evaluate the correctness of 50 CoT-answer pairs from six datasets and compare inconsistency ratios in them. The main results on Llama3.1-8B are illustrated in Table \ref{tab:paradigm_faith} (results on other models are presented in Appendix \ref{append:faith_eval}). We can conclude that: \textbf{ The logical reasoning tasks have more unfaithful CoT issues.} Compared to other datasets, the proportion of inconsistencies is higher in logical reasoning (7/50 in PW and 17/50 in PQA) and mainly consists of wrong CoTs leading to correct answers. Our research focuses on interpreting these unfaithful issues within logical reasoning datasets in the following sections. 

% We also repeat this experiment on other models (see Appendix \ref{sec:appendix}) and get the same conclusion.

\subsection{Question to CoT: Unfaithful CoT misses correct information from context}\label{sec:4.2}
We seek to explore why CoTs lack such correct information in unfaithful cases. Since CoTs are generated based on the question, we hypothesize that it is due to the lack of information from the context of the question. To demonstrate it, we use IG (see Eq.\ref{eq:ig}) to compare the information interaction between questions and CoTs. 

\paragraph{Experimental Setup}
We experiment with three settings: `unfaithful', `faithful', and `average'. For `unfaithful', we select all of the unfaithful samples, calculating $IG(Q,C)$. For `faithful', we select samples where both CoT and the answer are correct (see Figure \ref{fig:bridge} for examples). For `average', we calculate IG on all questions. We collect 200 samples from ProofWriter and ProntoQA, comparing the IG distribution under different settings.

\begin{figure}[tbp] 
    \centering
 %    \subfigbottomskip=2pt %两行子图之间的行间距
	% \subfigcapskip=-5pt %设置子图与子标题之间的距离
    \begin{subfigure}[t]{.49\linewidth}
        \centering
	\includegraphics[width=\linewidth]{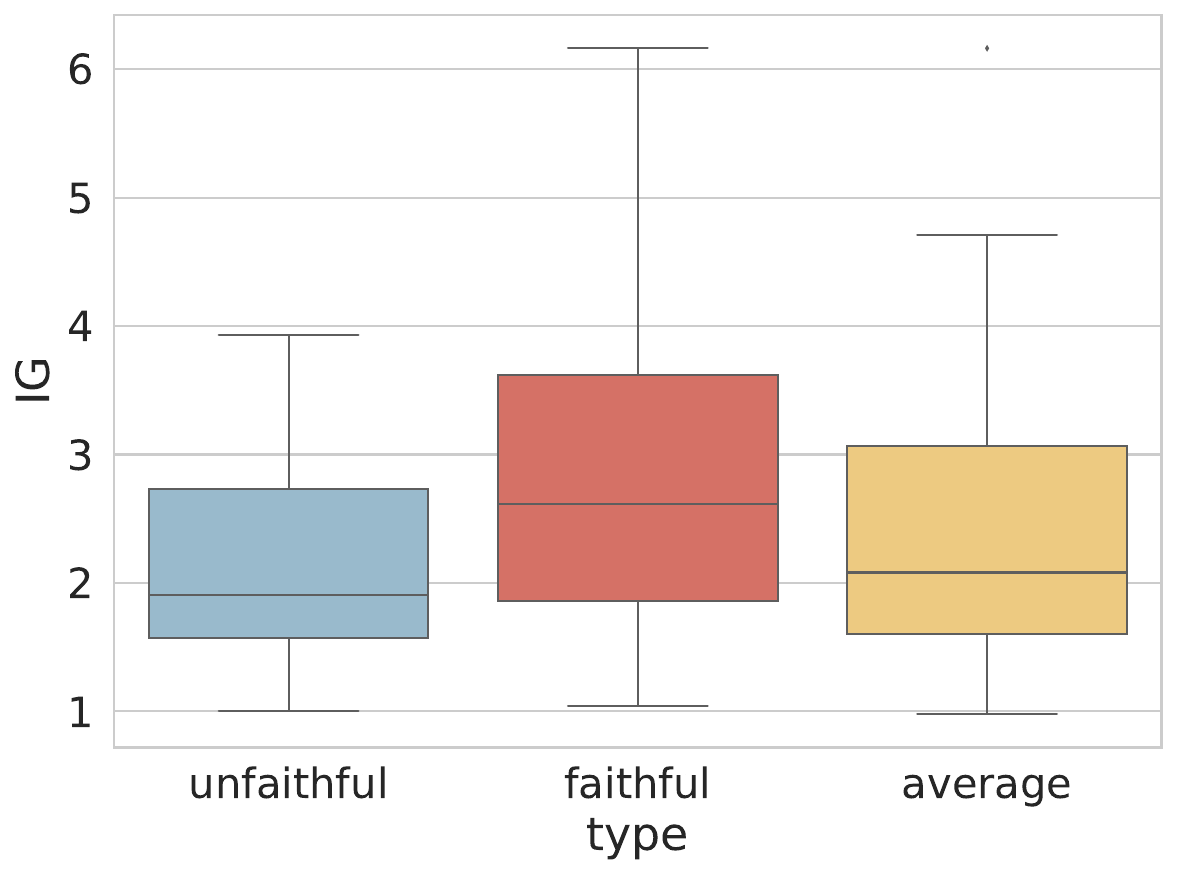}
        \caption{ProofWriter}
    \end{subfigure}
    \begin{subfigure}[t]{.49\linewidth}
        \centering
	\includegraphics[width=\linewidth]
    {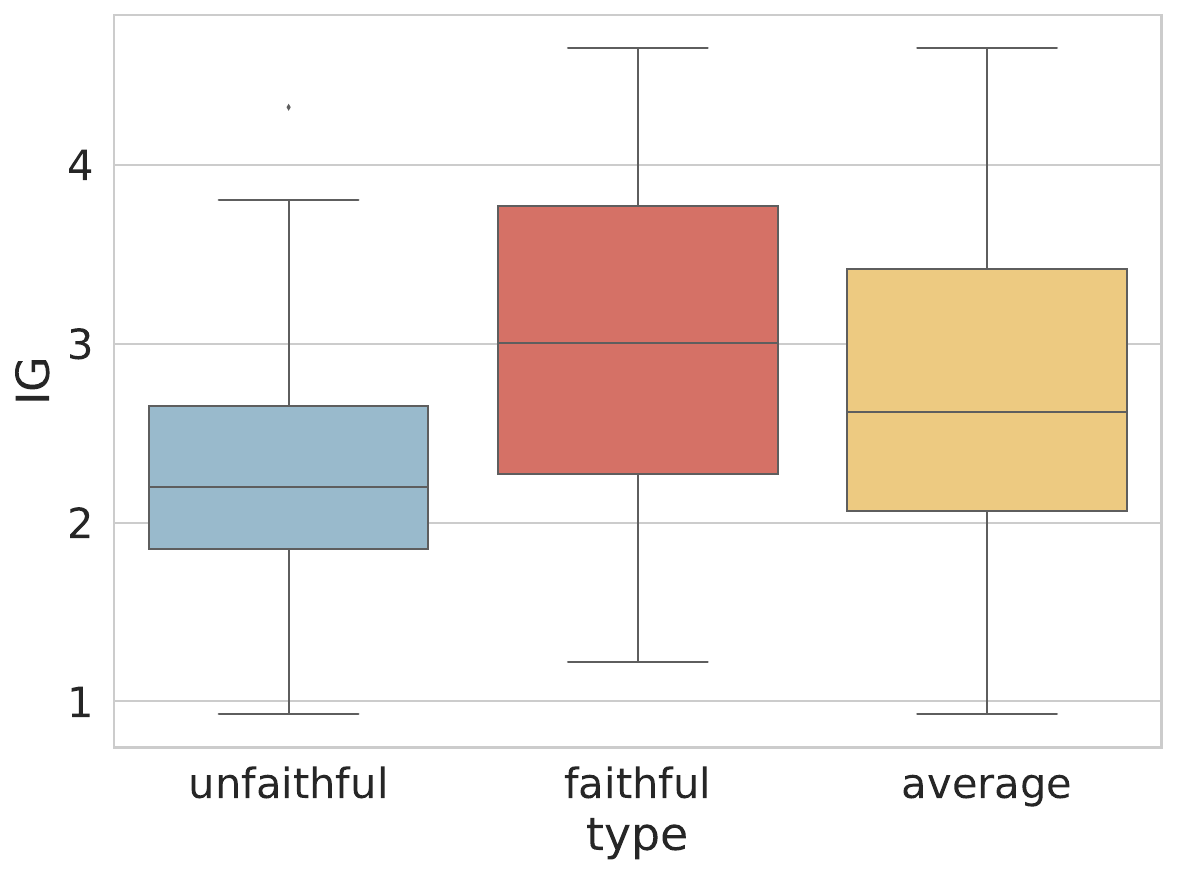}
        \caption{ProntoQA}
    \end{subfigure}
    \\
    \caption{Comparison of information transfer between questions and CoTs under three settings.}
    \label{fig:qc}
\end{figure}

\paragraph{Experimental Results}
Figure \ref{fig:qc} presents our results (we present more experiments in Appendix \ref{append:question2cot}). We can get that: \hypertarget{Cl.4}{}\textbf{(Cl.4) Unfaithful CoT misses correct information from the context.} In both figures, the IG under the `unfaithful' setting is lower than the other two settings. This indicates that CoT gets less information from the context when an unfaithful issue occurs. As an example, in unfaithful CoT of Figure \ref{fig:bridge}, the incorrect CoT does not contain the statement ``\textit{Gary is quiet}'' or ``\textit{All round, quiet things are not blue}'' in the question.

\subsection{CoT to Answer: Unfaithful CoT has less information transfer to answers }\label{sec:4.3}

Since unfaithful CoT lacks the correct information needed for reasoning, why can the final prediction still be correct? To answer it, we investigate the information transfer between CoT and the answer. 

\paragraph{Experimental Setup}
We use the AAE from the Eq.\ref{eq:aae} to measure the amount of information transferred between the two.
Following the experiment in $\S$\ref{sec:4.2}, we experiment under ``unfaithful'' and ``faithful'' settings, comparing AAE values on Llama3.1-8B across different datasets.
       
\begin{figure}[tbp] 
    \centering
 %    \subfigbottomskip=2pt %两行子图之间的行间距
	% \subfigcapskip=-5pt %设置子图与子标题之间的距离
    \begin{subfigure}[t]{.49\linewidth}
        \centering
	\includegraphics[width=\linewidth]{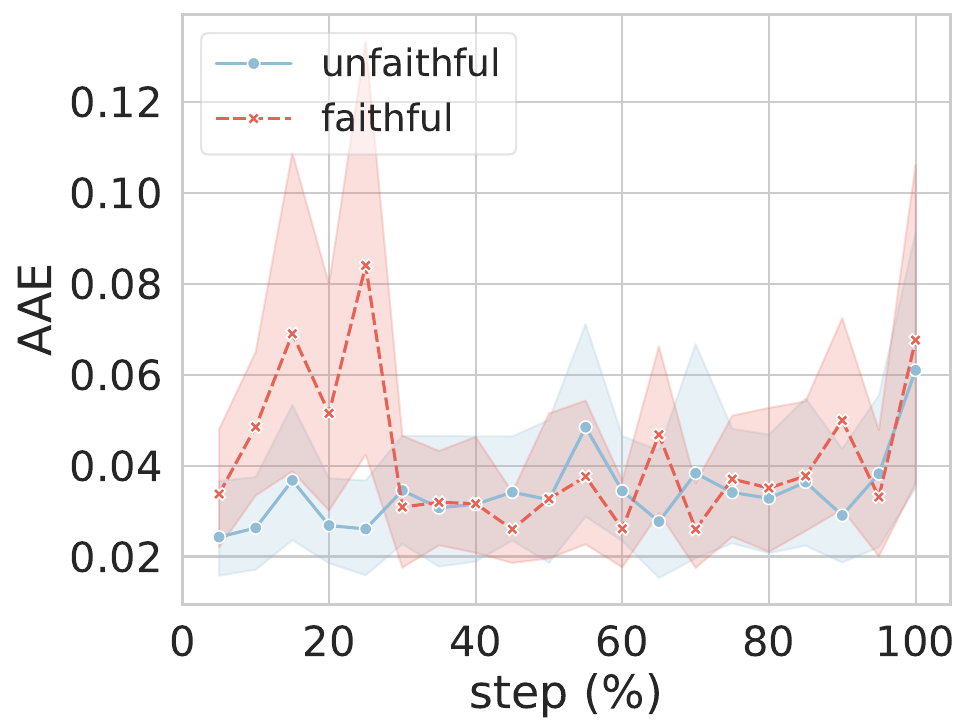}
        \caption{ProofWriter}
    \end{subfigure}
    \begin{subfigure}[t]{.49\linewidth}
        \centering
	\includegraphics[width=\linewidth]
    {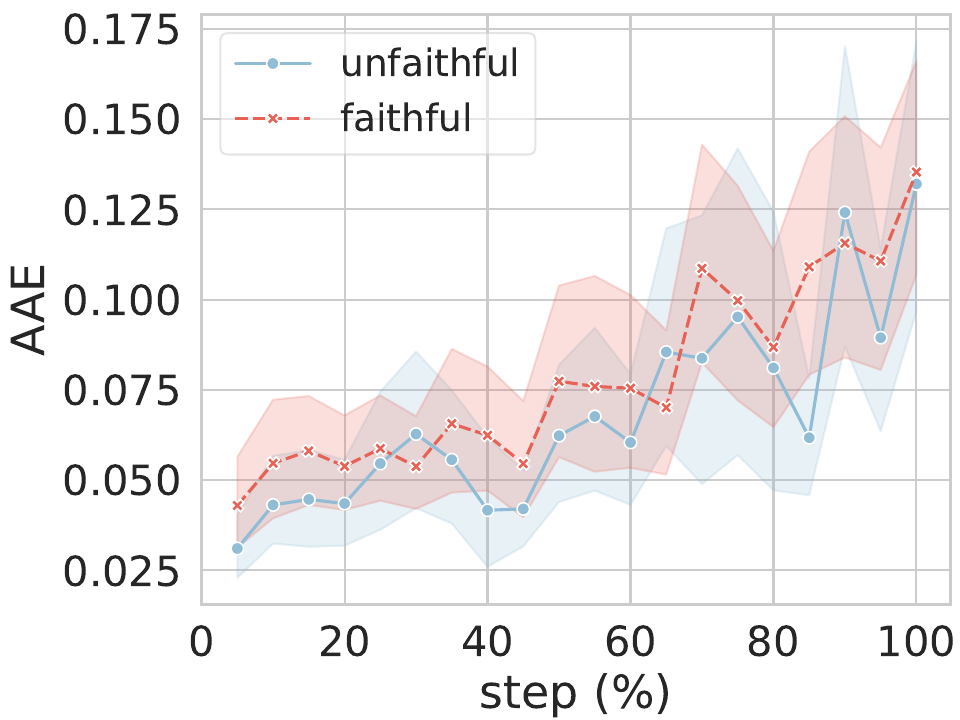}
        \caption{ProntoQA}
    \end{subfigure}
    \\
    \caption{Comparison of information transfer between CoTs and answers on Llama3.1-8B.}
    \label{fig:ca}
\end{figure}

\paragraph{Experimental Results}
The main results of the experiments are demonstrated in Figure \ref{fig:ca}. In both figures, the AAE for the `faithful' setting (in red) is higher than that for the `unfaithful' setting (in blue). Therefore, we have:
\hypertarget{Cl.5}{\textbf{(Cl. 5) Unfaithful CoT has less information interaction with the answer compared to the correct one.}}

% \begin{figure}[htbp] 
%     \centering
%  %    \subfigbottomskip=2pt %两行子图之间的行间距
% 	% \subfigcapskip=-5pt %设置子图与子标题之间的距离
%     \begin{subfigure}[t]{.245\linewidth}
%         \centering
% 	\includegraphics[width=\linewidth]{Styles/fig/Llama2_13b_pw_pans_f.pdf}
%         \caption{Centralized Reasoning}\label{fig:cent_reason}
%     \end{subfigure}
%     \begin{subfigure}[t]{.245\linewidth}
%         \centering
% 	\includegraphics[width=\linewidth]{Styles/fig/Llama2_13b_pw_pans_unf.pdf}
%         \caption{Distributed Reasoning}\label{fig:dist_reason}
%     \end{subfigure}
%      \begin{subfigure}[t]{.245\linewidth}
%         \centering
% 	\includegraphics[width=\linewidth]{Styles/fig/Llama2_13b_pqa_pans_f.pdf}
%         \caption{Centralized Reasoning}\label{fig:cent_reason}
%     \end{subfigure}
%     \begin{subfigure}[t]{.245\linewidth}
%         \centering
% 	\includegraphics[width=\linewidth]{Styles/fig/Llama2_13b_pqa_pans_unf.pdf}
%         \caption{Distributed Reasoning}\label{fig:dist_reason}
%     \end{subfigure}
%     \\
%     \caption{Two reasoning paradigms.}
%     \label{fig:cot_paradigm}
% \end{figure}

\subsection{Question to Answer: Answer can recall correct information from context}\label{sec:4.4}
While the answer misses key information from the CoT, how can the final prediction still be correct?  We hypothesize that LLMs can recall the missing information when generating the answer and design experiments to demonstrate it.

\paragraph{Experimental Setup}
We rank each statement in the context by its AAE score to the answer $AAE(S, A)$ ($S$ is a statement in the question) and observe whether the top-ranked statements include the correct statement missing in CoT (e.g. ``\textit{Gary is quiet}'' in Figure \ref{fig:bridge}). For comparison, we conduct experiments under three settings: unfaithful (unfaithful CoT with AAE recall), average (all CoT with AAE recall), and random (unfaithful CoT with random recall).

\begin{figure}[tbp] 
    \centering
 %    \subfigbottomskip=2pt %两行子图之间的行间距
	% \subfigcapskip=-5pt %设置子图与子标题之间的距离
    \begin{subfigure}[t]{.49\linewidth}
        \centering
	\includegraphics[width=\linewidth]{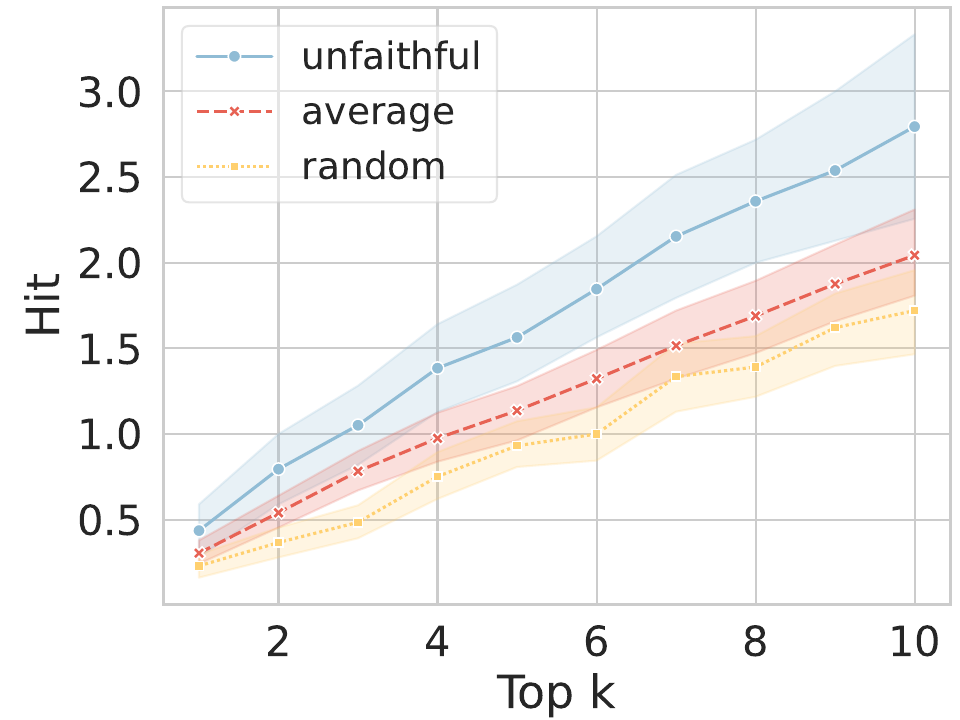}
        \caption{ProofWriter}
    \end{subfigure}
    \begin{subfigure}[t]{.49\linewidth}
        \centering
	\includegraphics[width=\linewidth]
    {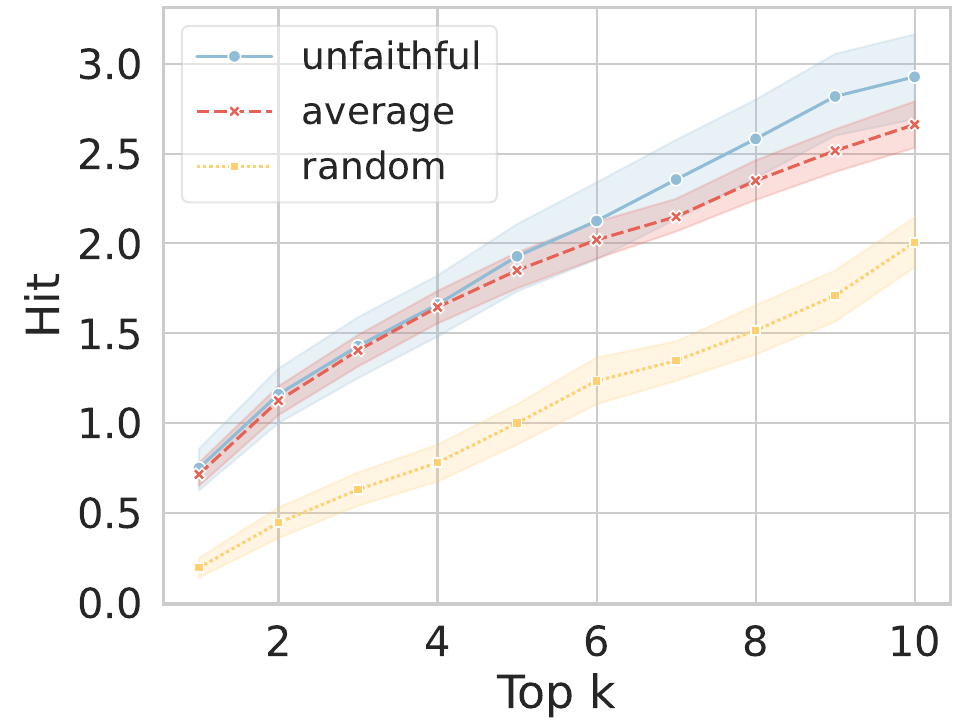}
        \caption{ProntoQA}
    \end{subfigure}
    \\
    \caption{Comparison of correct recall counts.}
    \label{fig:context_answer}
\end{figure}

\begin{figure*}[tbp] 
    \centering
	\includegraphics[width=0.9\linewidth]{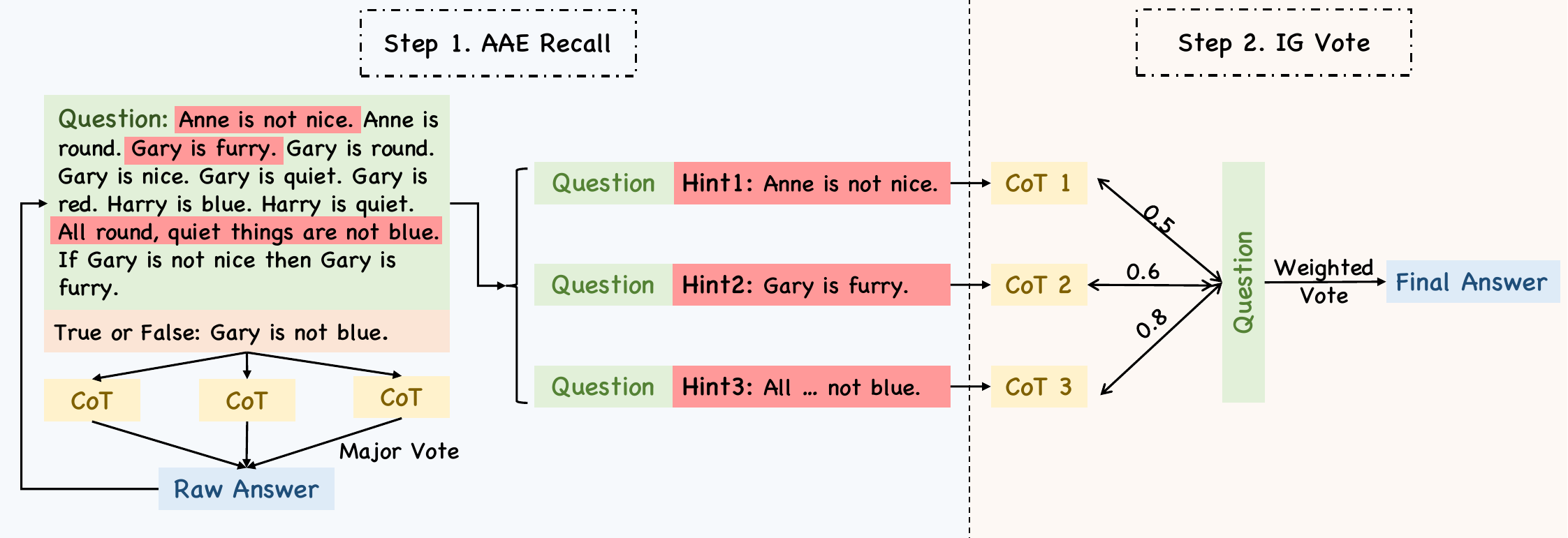}
    \caption{The main process of our QUIRE method, where the statement in red is the recalled information.}
    \label{fig:method}
\end{figure*}
\paragraph{Experimental Results}
Figure \ref{fig:context_answer} demonstrates our results on Llama3.1-8B (results on more models in Appendix \ref{append:question2answer}), from which we conclude that: \hypertarget{Cl.6}{}\textbf{(Cl. 6) When unfaithful CoT issues occur, LLMs can recall missing correct information from the context during the answer prediction.} For all datasets and models, when the unfaithful CoT issue occurs, more missing statements get the top-k highest AAE scores from the answer compared to other settings. These statements have a strong information interaction with the answer, compensating for the lack of relevant statements in the CoT, thereby contributing to the correct answer prediction. 

% \textbf{This effect, combined with the findings from \hyperlink{Cl.4}{Cl.4} and \hyperlink{Cl.5}{Cl.5}, leads to inconsistencies in the context information obtained by the CoT and the answer, resulting in the ineffectiveness.}

\section{From Unfaithful CoT to Effective CoT}\label{sec:5}
Since we analyze the CoT from two different perspectives in the former experiments, what is the relationship between them? In this section, we demonstrate that mitigating the unfaithful issue can lead to improvements in final performance. In other words, the faithfulness of CoT ($\S$\ref{sec:4}) is a key factor in influencing the CoT effectiveness ($\S$\ref{sec:3}).

\subsection{Our Method}\label{sec:5.1}
Based on findings in $\S$\ref{sec:4}, we propose a new method called \textbf{QU}estion \textbf{I}nformation \textbf{R}ecall and \textbf{E}nhancement (QUIRE) to mitigate the unfaithful CoT issue.
The main framework of it is illustrated in Figure \ref{fig:method}, which includes two components:
\paragraph{AAE Recall} As mentioned in \hyperlink{Cl.6}{Cl.6}, when unfaithful issues occur, LLMs maintain a strong causal relevance with the correct statement in the context during the answer prediction. 
Thus, here we first generate a raw answer $A$ with the Self-Consistency (SC) method, then recall extra information by selecting the top-k context statements with the highest $AAE(S, A)$ (as marked with red in Figure \ref{fig:method}). After recalling extra information, we incorporate these statements as additional hints into the input prompt, enabling the model to pay more attention to this information during the CoT generation.

\paragraph{IG Vote} Through the former step, we get multiple information-enhanced CoTs (here we can also integrate the SC technique to further improve the performance). However, since our recall method also introduces noisy hints, there may exist incorrect statements in some of these CoTs (e.g. Hint 1 in Figure \ref{fig:method}). To reduce their interference, according to \hyperlink{Cl.4}{Cl.4}, we rate these CoTs based on $IG(Q,C)$. A higher IG indicates that more information in CoT is derived from the question, which means the CoT contains fewer hallucinated statements. After calculation, we use these scores as the weight for SC to vote and select the final answer.

\subsection{Main Experimental Setup}\label{sec:5.2} 
\paragraph{Datasets} Since all analyses in $\S$\ref{sec:4} are conducted on ProofWriter \citep{proofwriter} ProntoQA \citep{prontoqa}, we continue to evaluate our method on them. For the test set, we sample 500 questions from the former and 400 questions from the latter.

\paragraph{Metrics} In form sections, we analyze the CoT performance from two aspects. Therefore, our evaluation cannot solely consider the result performance but should also assess the quality of the CoT to avoid unfaithful reasoning. Therefore, in addition to accuracy (Acc), we use the following two metrics: \textbf{(1) BertScore (BS):} Given a golden rationale, the generated CoT should recall as much information from it as possible, hence, we use the BertScore \citep{bertscore} as one of our metrics. \textbf{(2) Faithful BertScore (FBS):} From the perspective of faithfulness, correct answers should be accompanied by high-quality CoTs, and incorrect results should correspond to CoTs of poorer quality. Thus, we define the FBS to measure faithfulness as below:
    \begin{equation}\label{eq:information}
   \footnotesize{
   \begin{aligned}
FBS &= \frac{1}{n}\sum_{i=1}^{n} [\eta(a_i) BS(c_i,g_i) \\ & + (1-\eta(a_i)) (1 - BS(c_i,g_i)) ]
    \end{aligned}}
\end{equation}
where $c_i$, $a_i$, $g_i$ represent the generated CoTs, answers and golden rationales, $n$ denotes the sample count. If $a_i$ is correct, $\eta(a_i) = 1$, else $\eta(a_i) = 0$.
\paragraph{Baselines} For baselines, we select representative methods that enhance LLMs' reasoning performances, including: \textbf{Chain-of-Thought (CoT)} \citep{cot}, \textbf{Self-Consistency (SC)} \citep{cot-sc}, \textbf{Least-to-Most (LtM)} \citep{l2m}, \textbf{Self-Refine (SR)} \citep{sr}. 
Additionally, we also set up ablation experiments (-AAE Recall and -IG Vote) to verify the effectiveness of each component in our method. Implementation details can be found in Appendix \ref{append:main_exp}.

% We choose \texttt{Llama2-13b-base},\texttt{Mistral-7b-v0.1} as the models and use 4 NVIDIA A100 GPUs to run experiments.

\begin{table}[tbp]
\centering
 \scalebox{0.8}{
\begin{tabular}{lcccccc}
\toprule
\multirow{2}{*}{\textbf{Method}} & \multicolumn{3}{c}{\textbf{ProofWriter}} & \multicolumn{3}{c}{\textbf{ProntoQA}} \\
\cmidrule(r){2-4} \cmidrule(r){5-7}
& Acc & BS & FBS & Acc & BS & FBS   \\
\midrule
 CoT & 59.2  & 64.9 & 55.7 & 86.8 & 86.1 & 78.0 \\ 
 SC & 60.6  & 65.0 & 57.8 & 93.2 & 87.5 & 83.6 \\ 
% SC@10 & 69.6  & 42.4 & 47.0 & 96.0 & 61.2 & 61.6 \\ 
LtM  & 54.0 & 60.4 & 56.4 & 90.0 & 77.3 & 72.6 \\
 SR  & 51.6 & 65.9 & 53.4 & 88.5 & 91.5 & 84.5\\
\midrule
\textbf{Ours} & \textbf{63.0} & \textbf{66.6} & \textbf{58.0} & \textbf{95.0} & \textbf{92.7} & \textbf{89.2}  \\
 - AAE Recall & 60.2 & 65.1 & 57.0 & 95.0 & 87.5 & 84.6\\
 - IG Vote & 62.8 & 64.1 & 56.6 & 94.3 & 87.0 & 83.4\\ 
\bottomrule

\end{tabular}}
\caption{Results of our main experiment, the best results are highlighted in \textbf{bold}.}
\label{tab:main}
\end{table}

\subsection{Main Experimental Results}\label{sec:5.3} 
The results of our main experiment on Llama3.1-8B are demonstrated in Table \ref{tab:main} (additional results in Appendix \ref{append:main_result}), which demonstrates that: \textbf{(1) Our method effectively mitigates the unfaithful CoT issues.} On both BS and FBS, our method achieves the highest performance, improving up to \textbf{5.6\%} faithfulness (i.e. FBS) on ProntoQA. Besides, from the results of the ablation study, we can see both modules make contributions to enhancing the CoT faithfulness. Given that our method is an application derived from the analytical conclusions, its superior performance can also substantiate the correctness of our earlier findings.
\textbf{(2) Improvements in faithfulness can also lead to enhancements in CoT's effectiveness.} Although our method is based on the conclusions from \S\ref{sec:4} to optimize the unfaithful CoT issue, the CoT effectiveness (Acc) also improved (up to \textbf{2.4\%} on ProofWriter), indicating that the former is a significant factor influencing the latter. Through our method, we can boost the CoT's performance from both effectiveness and faithfulness.

% \begin{wrapfigure}{b}{0.5\linewidth}
% \vspace{-3mm}
    
% \vspace{-6mm}
% \end{wrapfigure}

\section{Conclusion}\label{sec:6}
In this paper, we focus on analyzing the CoT performance in reasoning tasks. Specifically, we identify the factors influencing CoT effectiveness and interpret the mechanism behind CoT unfaithfulness. For the former, we conduct extensive experiments to demonstrate that question difficulty, information gain, and information flow all contribute to CoT's performance improvement. For the latter, we capture the information transfer among questions, CoTs, and answers in the reasoning process. The experimental results indicate that the information recall mechanism during answer predictions leads to unfaithful CoT issues. At last, we design the QUIRE method as a preliminary application of our findings, which significantly improves CoT performances from both perspectives.

\section*{Limitations}
Although our work conducts an in-depth analysis and proposes mitigation strategies for improving CoT performance, it has several limitations.
Firstly, due to the inability to access gradient information inside models like GPT-4, our analysis is limited to open-source LLMs. 
Secondly, although we have empirically demonstrated that improvements in faithfulness can lead to performance enhancements, there is still a lack of corresponding theoretical proof to support this conclusion.
We leave the CoT effectiveness analysis of black-box LLMs and further theoretical proof for our future work.

\section*{Acknowledgments}
This work is supported by the National Natural Science Foundation of China (No. U24A20335, No. 62176257, No. 62406321). This work is also supported by the Youth Innovation Promotion Association CAS and the China Postdoctoral Science Foundation under Grant Number 2024M753500. 

\bibliography{custom}
\newpage
\appendix

\section{Additional Experiments across Different Difficulty Levels}
\label{append:diff}
\begin{figure*}[htbp] 
    \centering
	\includegraphics[width=0.9\linewidth]{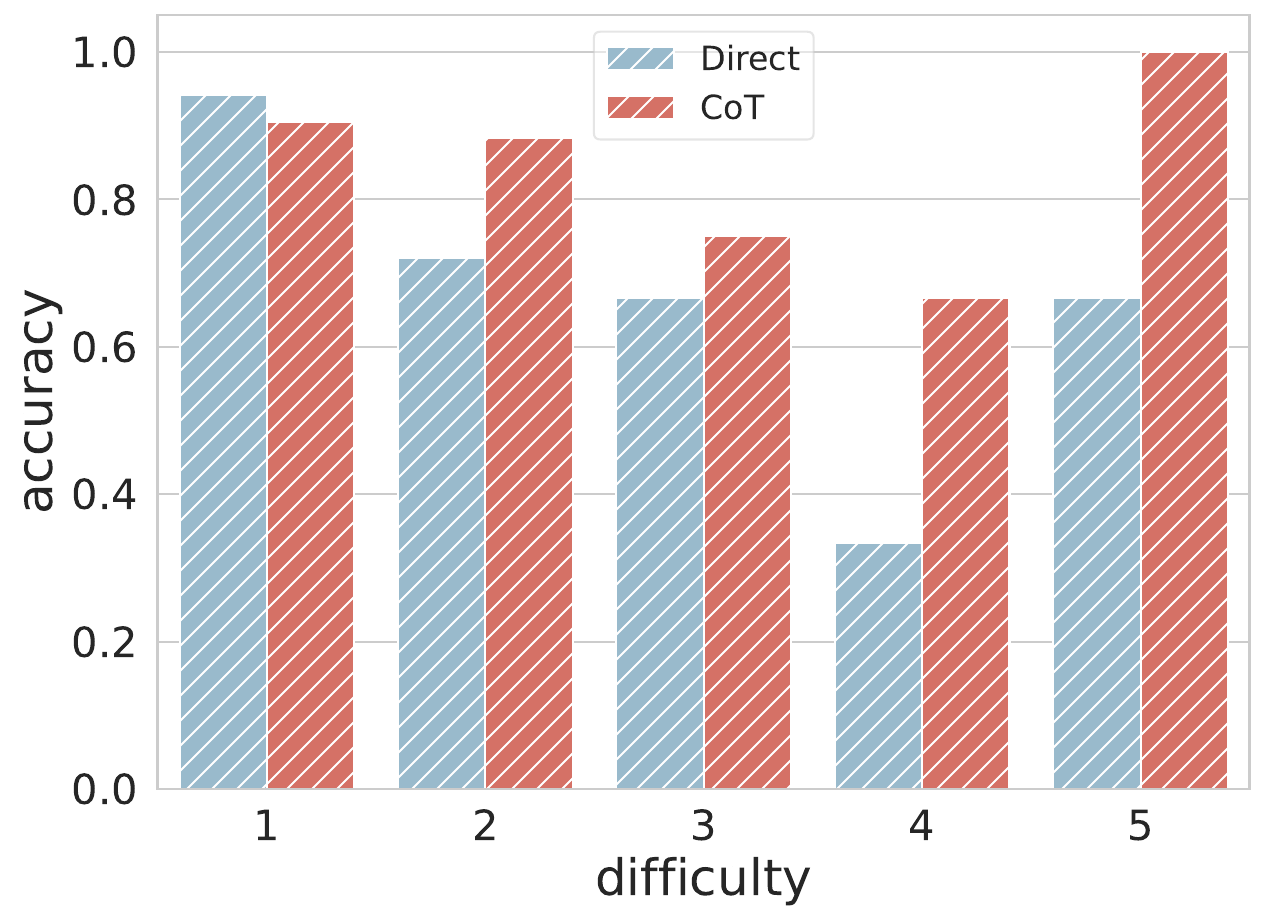}
    \caption{Performance on different problem difficulty
levels with and without CoT prompting (Llama3.1-8B on ProntoQA). }
    \label{fig:prontoqa_diff}
\end{figure*}

\begin{figure*}[htbp] 
    \centering
	\includegraphics[width=0.9\linewidth]{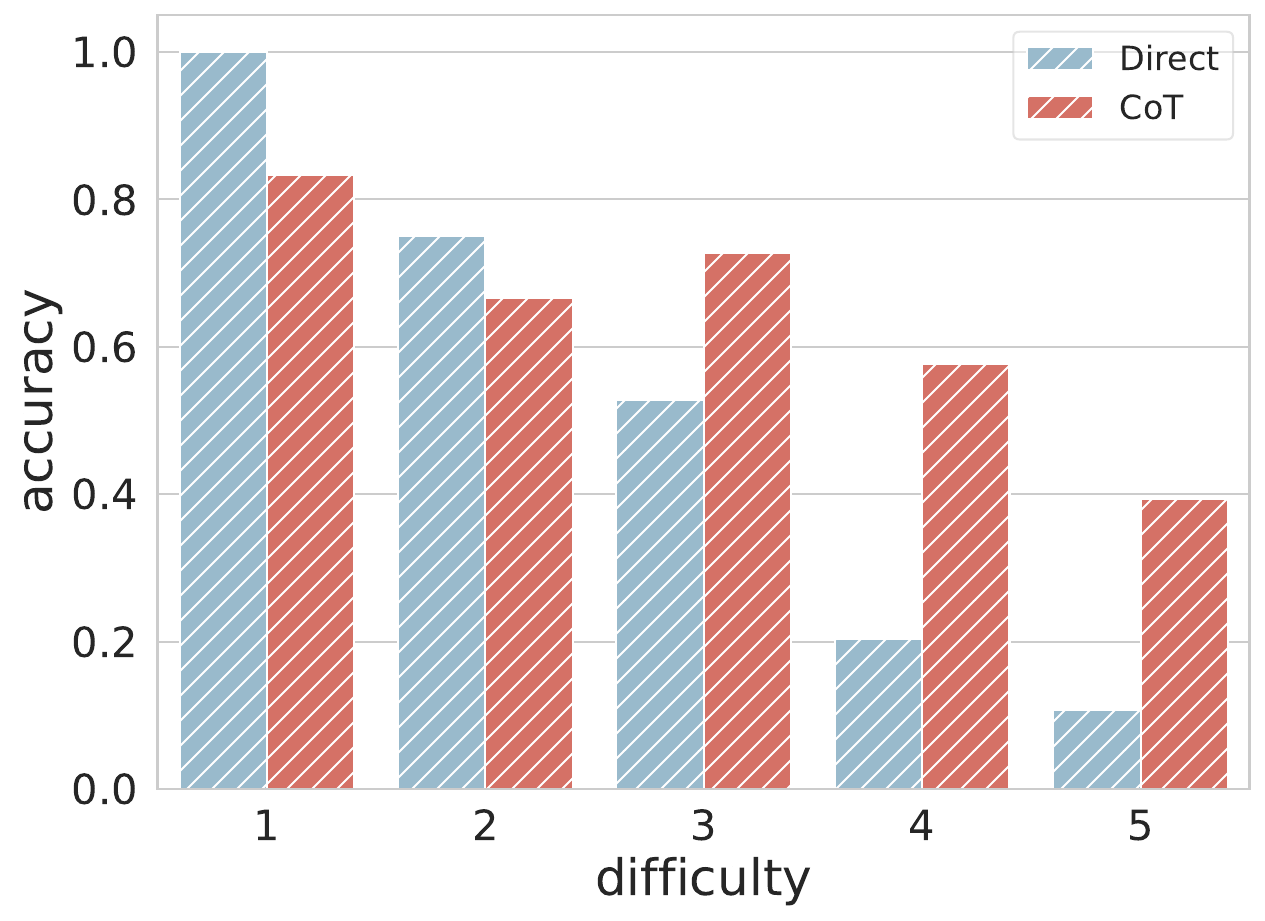}
    \caption{Performance on different problem difficulty
levels with and without CoT prompting (Gemma2-9B on AQuA). }
    \label{fig:aqua_diff}
\end{figure*}

\begin{figure*}[htbp] 
    \centering
	\includegraphics[width=0.9\linewidth]{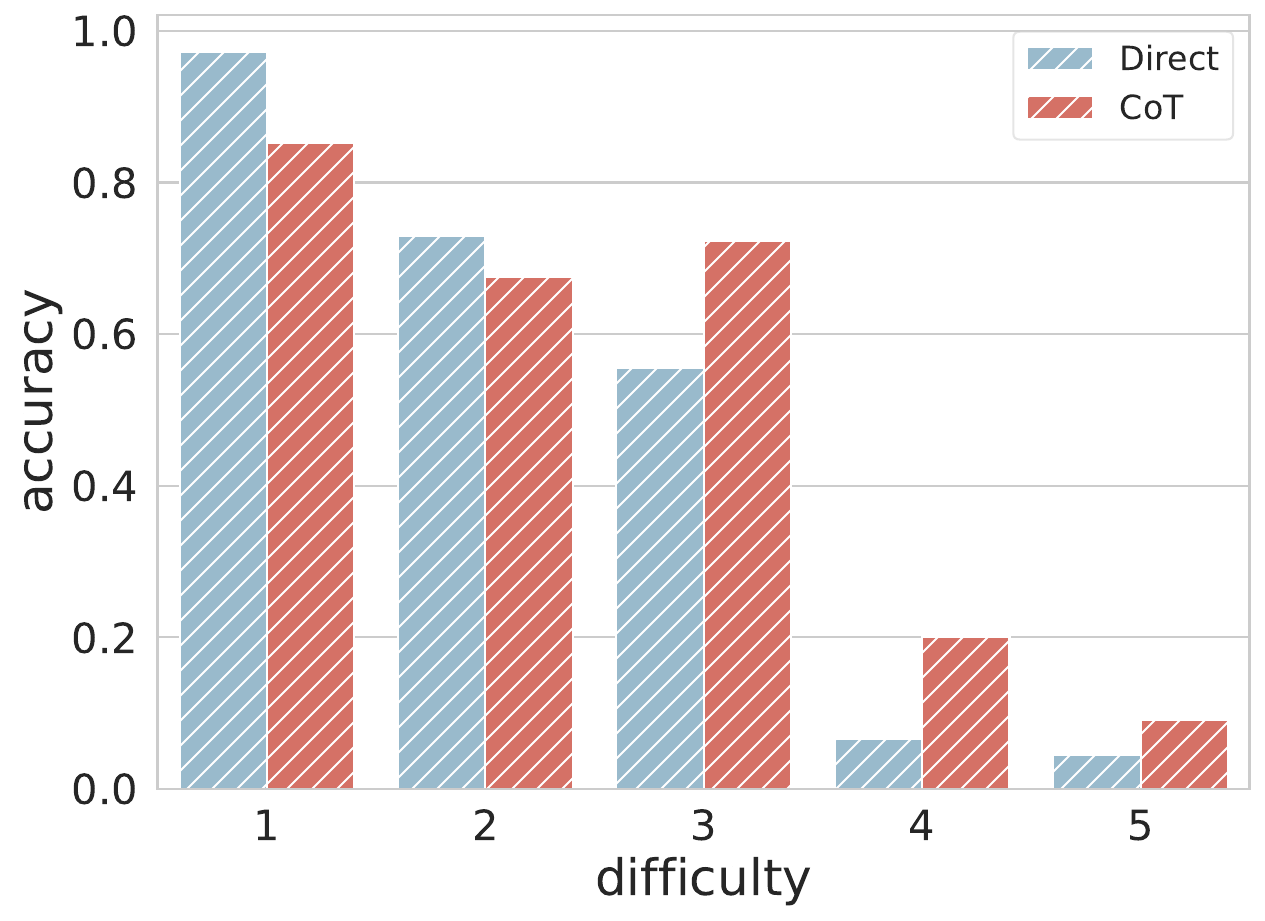}
    \caption{Performance on different problem difficulty
levels with and without CoT prompting (Gemma2-9B on SIQA). }
    \label{fig:siqa_diff}
\end{figure*}

\begin{figure*}[htbp] 
    \centering
	\includegraphics[width=0.9\linewidth]{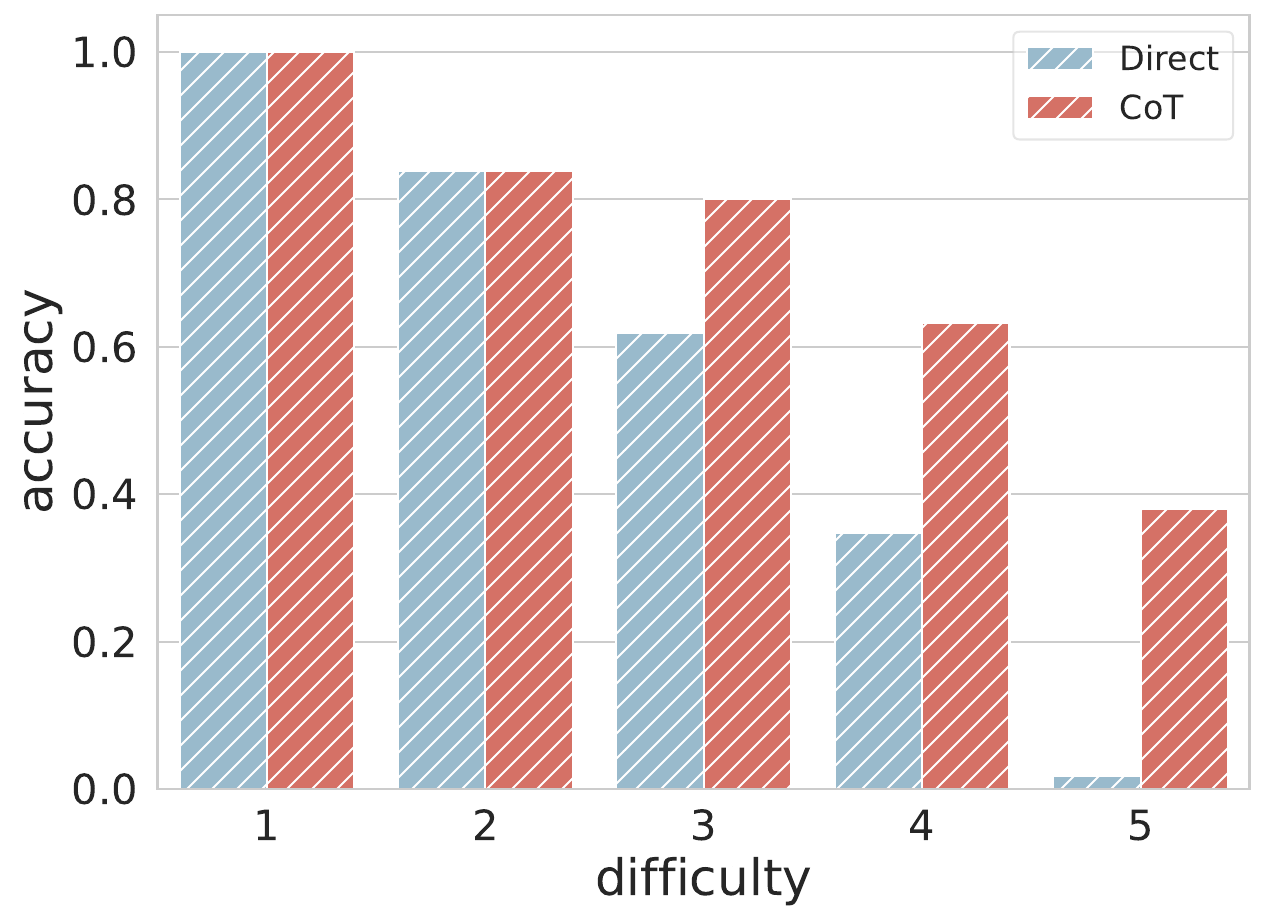}
    \caption{Performance on different problem difficulty
levels with and without CoT prompting (Gemma2-9B on ProofWriter). }
    \label{fig:proofwriter_diff}
\end{figure*}

\begin{figure*}[htbp] 
    \centering
	\includegraphics[width=\linewidth]{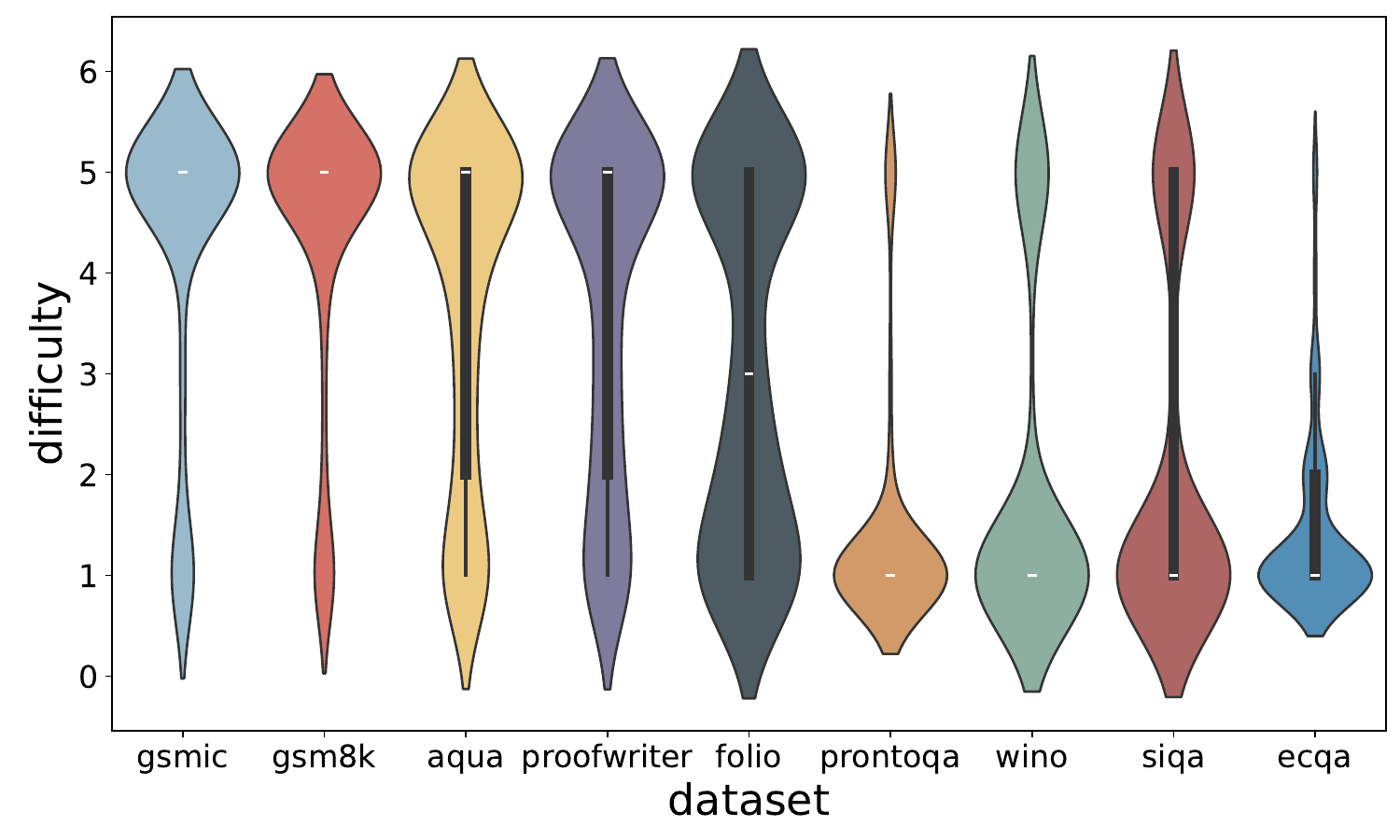}
    \caption{Difficulty distribution in different datasets on Gemma2-9B. }
    \label{fig:gemma_diff}
\end{figure*}

\begin{figure*}[htbp] 
    \centering
\includegraphics[width=\linewidth]{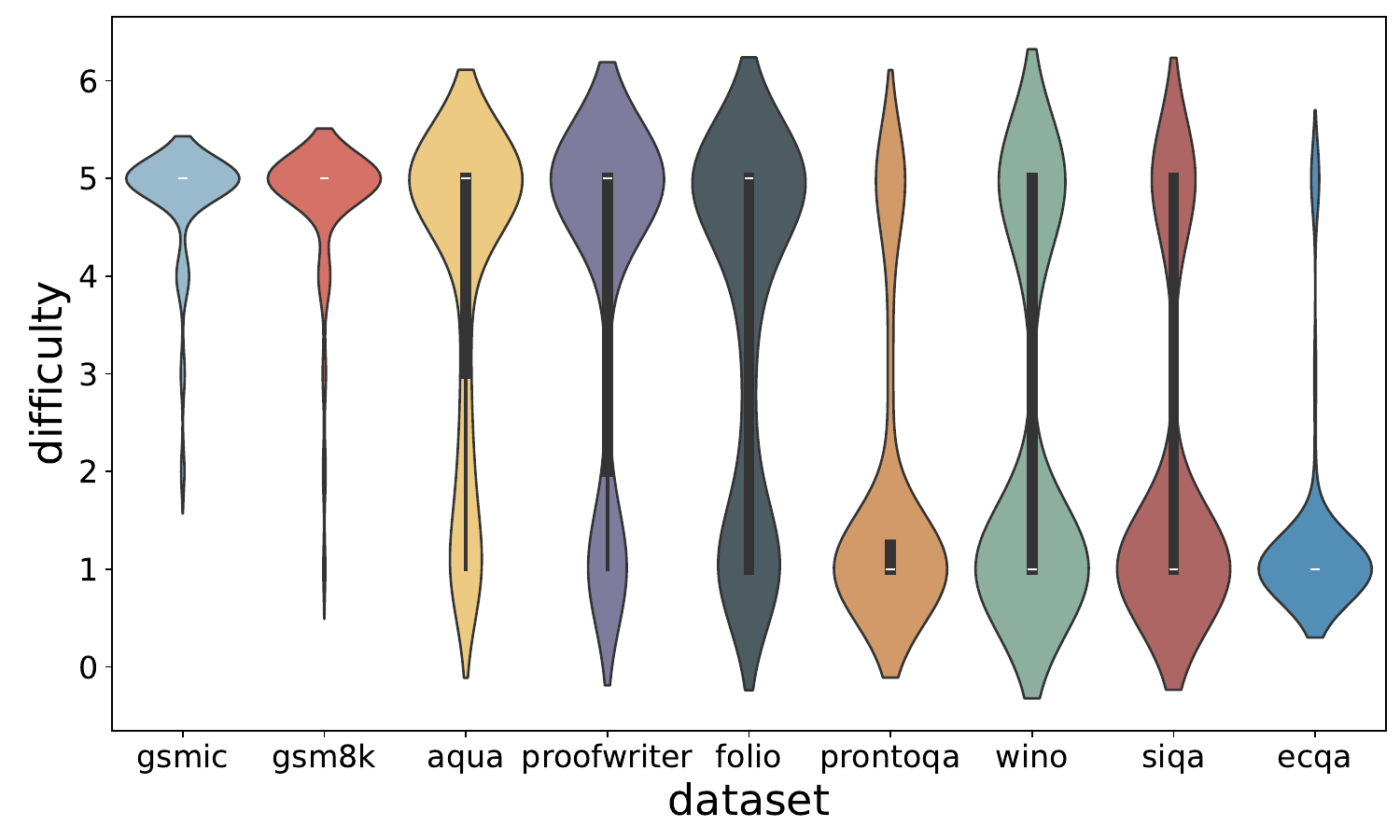}
    \caption{Difficulty distribution in different datasets on Mistral-7B. }
    \label{fig:mistral_diff}
\end{figure*}

In the main text, due to space constraints, we only presented results on GSM8k and WinoGrande, here we show more results on other datasets and models in Figure \ref{fig:prontoqa_diff}, \ref{fig:aqua_diff}, \ref{fig:siqa_diff}, \ref{fig:proofwriter_diff}. Besides, we also show the difficulty distribution on more models in Figure \ref{fig:gemma_diff} and \ref{fig:mistral_diff}.
These results are consistent with our conclusions in \hyperlink{Cl.1}{Cl.1}. 

\section{Details and Additional Experiments on Faithfulness Evaluation} \label{append:faith_eval}
To demonstrate the widespread existence of unfaithful issues in logical tasks, we present the evaluation results on Llama2-13B in Table \ref{tab:llama2_faith}.

\section{Additional Experiments on Question to CoT Information Analysis}\label{append:question2cot}
\begin{figure*}[htbp] 
    \centering
 %    \subfigbottomskip=2pt %两行子图之间的行间距
	% \subfigcapskip=-5pt %设置子图与子标题之间的距离
    \begin{subfigure}[t]{.49\linewidth}
        \centering
	\includegraphics[width=\linewidth]{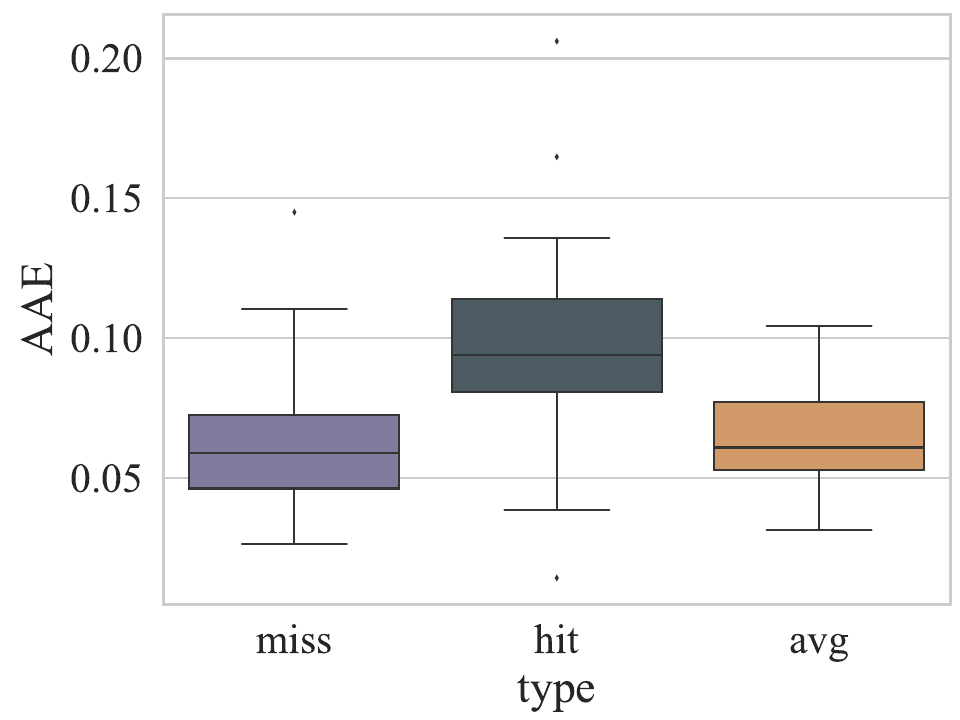}
        \caption{ProofWriter}\label{fig:cent_reason}
    \end{subfigure}
    \begin{subfigure}[t]{.49\linewidth}
        \centering
	\includegraphics[width=\linewidth]{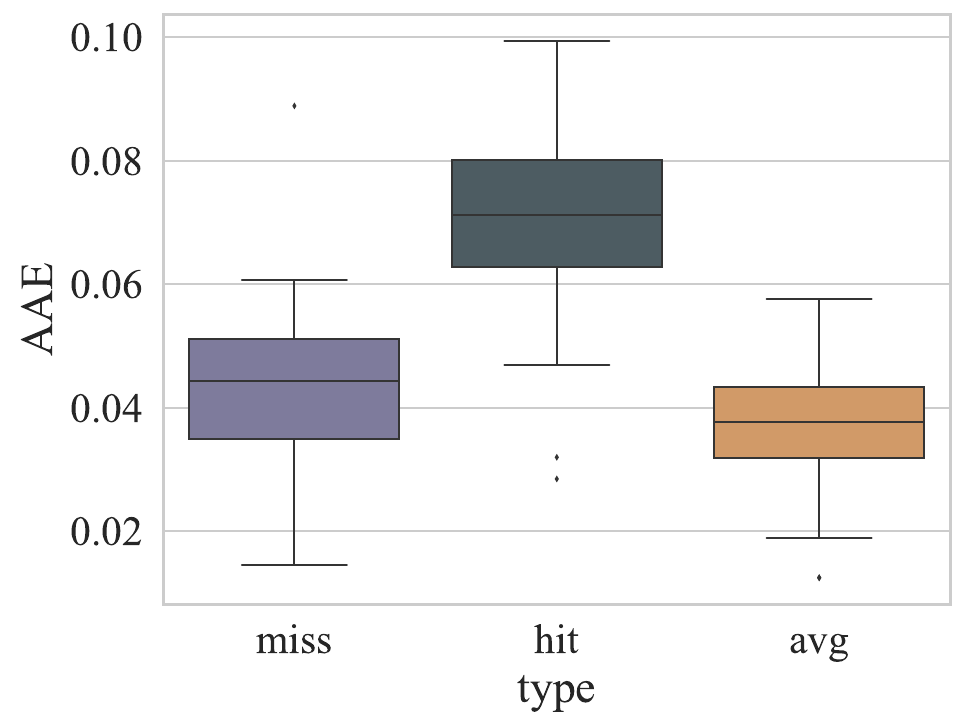}
        \caption{ProntoQA}\label{fig:dist_reason}
    \end{subfigure}
    \\
    \caption{Comparison of information interaction between contexts and CoTs under three settings (Llama2-13B).}
    \label{fig:question_cot_llama2}
\end{figure*}

\begin{figure*}[htbp] 
    \centering
 %    \subfigbottomskip=2pt %两行子图之间的行间距
	% \subfigcapskip=-5pt %设置子图与子标题之间的距离
     \begin{subfigure}[t]{.49\linewidth}
        \centering
	\includegraphics[width=\linewidth]{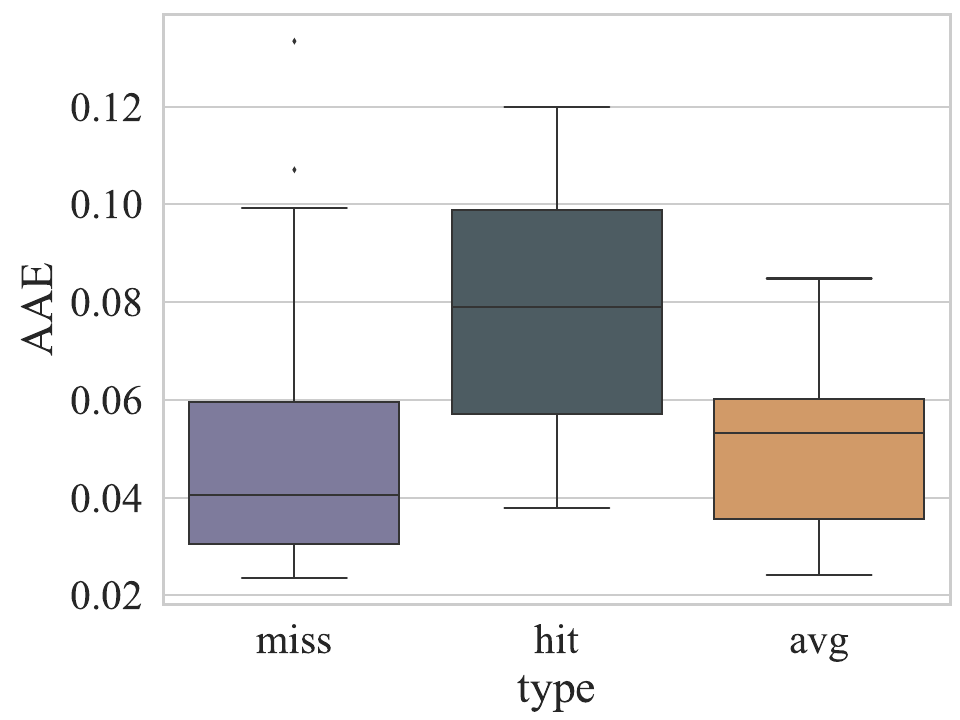}
        \caption{ProofWriter}
    \end{subfigure}
    \begin{subfigure}[t]{.49\linewidth}
        \centering
	\includegraphics[width=\linewidth]{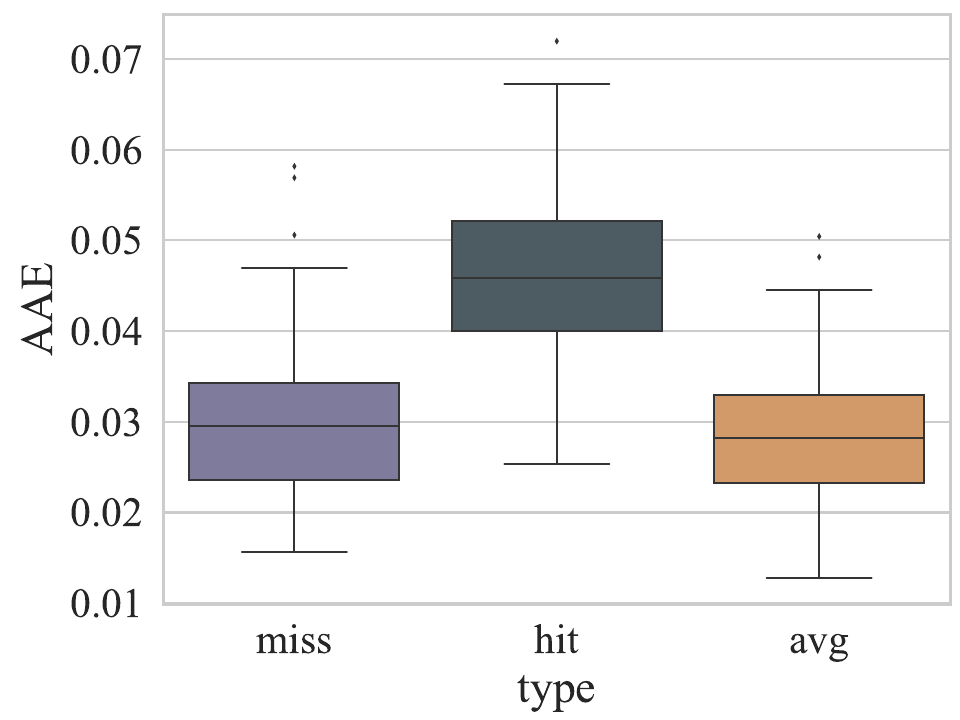}
        \caption{ProntoQA}
    \end{subfigure}
    \\
    \caption{Comparison of information interaction between contexts and CoTs under three settings (Mistral-7B).}
    \label{fig:question_cot_mistral}
\end{figure*}

In addition to the main experiments in \S\ref{sec:4.2}, here we conduct another experiment to further demonstrate our conclusion in \hyperlink{Cl.4}{Cl.4}. Specifically, we experiment with three settings: `miss', `hit', and `avg'. For `miss', we select the context statements that are present in the golden CoT (as provided in the dataset) but not in the generated CoT, calculating their $AAE(Q,C)$ scores to CoT. For  `hit', we collect the statements present in the generated CoT and compute the corresponding $AAE(Q,C)$. As for `avg', we calculate the $AAE(Q,C)$ between the whole context and CoT. We compare the distribution of the above three AAE scores on ProofWriter and ProntoQA (100 samples each) and illustrate the results in Figure \ref{fig:question_cot_llama2}, \ref{fig:question_cot_mistral}. Across all figures, the AAE for the `hit' setting is higher than that for the `miss' setting. Thus, compared to the question information present in the CoT, this missing information gets less attention from the model during the CoT generation. Besides, the score difference between the `hit' and the `avg' is also large, which means that the included context statements have a stronger information interaction with the CoT. The model tends to copy this attended information into the CoT. Therefore, the results are consistent with our findings in \S\ref{sec:4.2}.

\section{Additional Experiments on Question to Answer Information Analysis} \label{append:question2answer}
\begin{figure*}[htbp] 
    \centering
 %    \subfigbottomskip=2pt %两行子图之间的行间距
	% \subfigcapskip=-5pt %设置子图与子标题之间的距离
    \begin{subfigure}[t]{.49\linewidth}
        \centering
	\includegraphics[width=\linewidth]{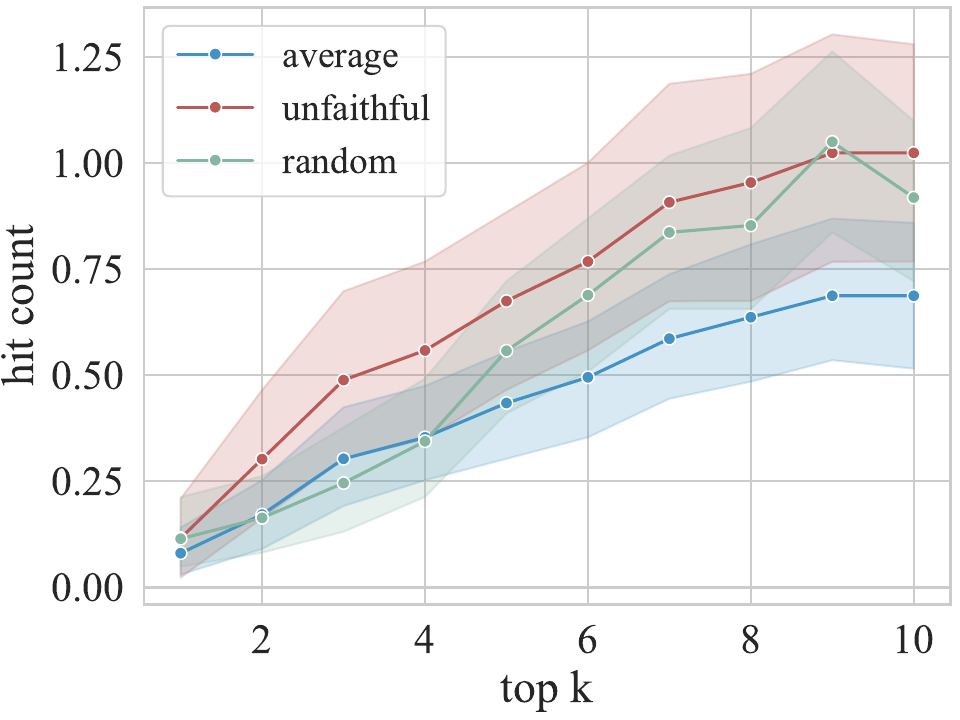}
        \caption{ProofWriter}\label{fig:cent_reason}
    \end{subfigure}
    \begin{subfigure}[t]{.49\linewidth}
        \centering
	\includegraphics[width=\linewidth]{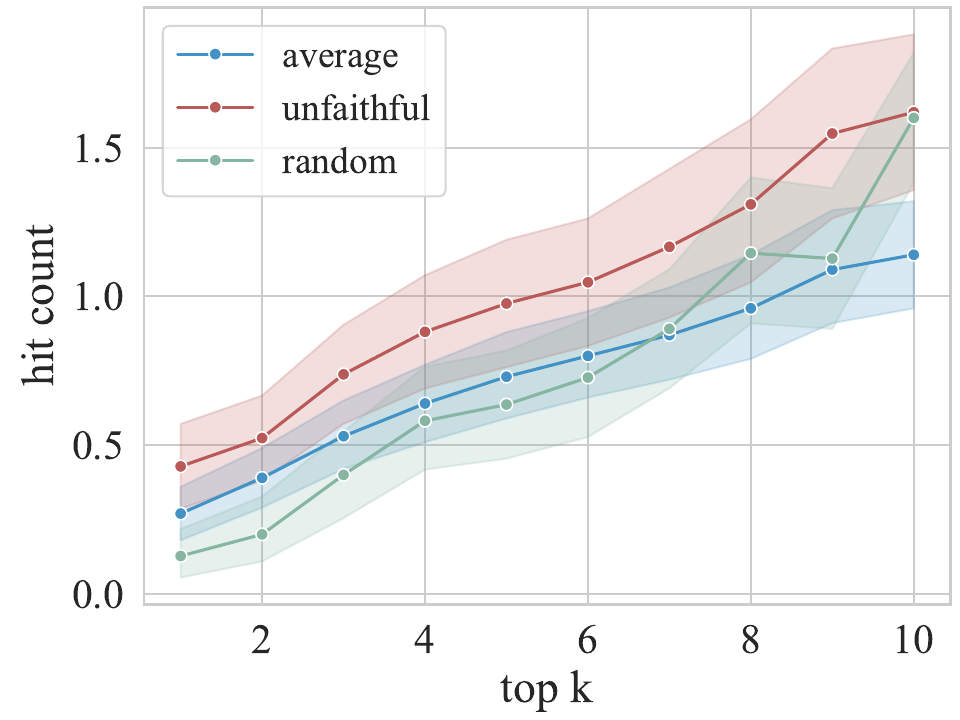}
        \caption{ProntoQA}\label{fig:dist_reason}
    \end{subfigure}
    \\
    \caption{Comparison of correct statements recall counts (Llama2-13B).}
    \label{fig:question_answer_llama2}
\end{figure*}

\begin{figure*}[htbp] 
    \centering
    
 %    \subfigbottomskip=2pt %两行子图之间的行间距
	% \subfigcapskip=-5pt %设置子图与子标题之间的距离
    \begin{subfigure}[t]{.49\linewidth}
        \centering
	\includegraphics[width=\linewidth]{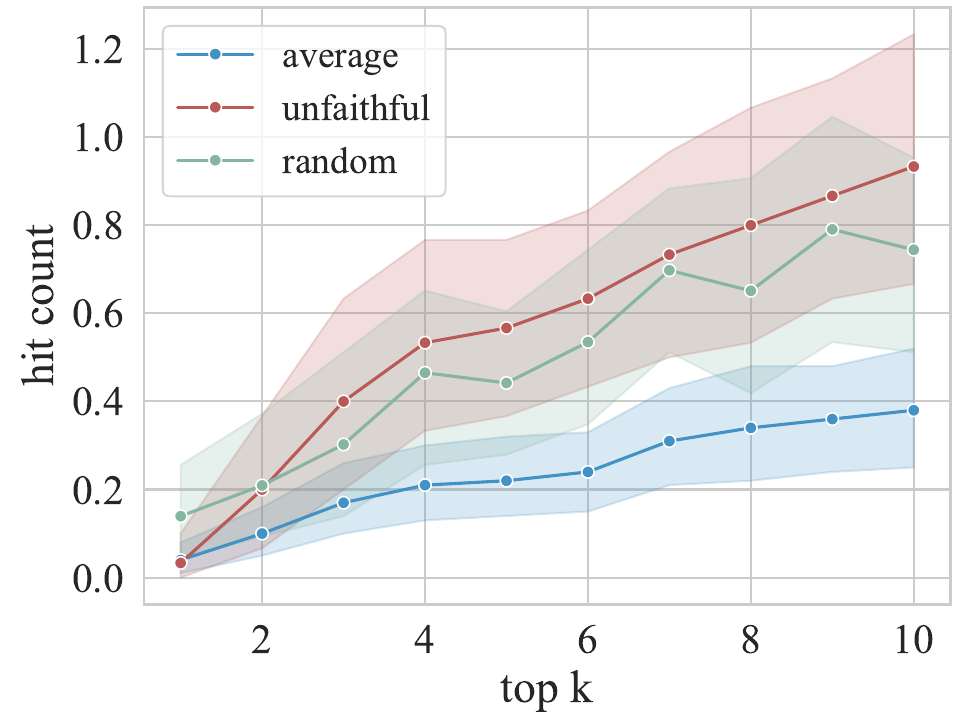}
        \caption{ProofWriter}\label{fig:cent_reason}
    \end{subfigure}
    \begin{subfigure}[t]{.49\linewidth}
        \centering
	\includegraphics[width=\linewidth]{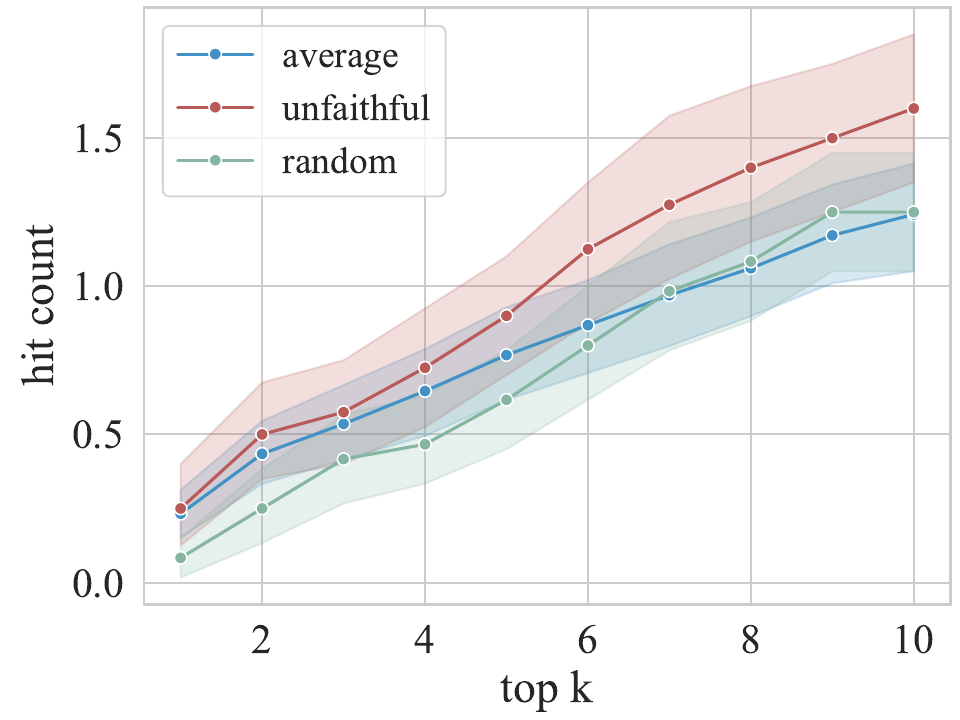}
        \caption{ProntoQA}\label{fig:dist_reason}
    \end{subfigure}
    \\
    \caption{Comparison of correct statements recall counts (Mistral-7B).}
    \label{fig:question_answer_mistral}
\end{figure*}

To demonstrate the generalizability of our conclusions in \S\ref{sec:4.4}, we repeat the experiments on two more models and present the result in Figure \ref{fig:question_answer_llama2} and \ref{fig:question_answer_mistral} (here we sample 100 questions from ProntoQA and ProofWriter). The results are consistent with \hyperlink{Cl.6}{Cl.6}.

\section{Implementation Details of the Main Experiment} \label{append:main_exp}
Here we provide a detailed account of the implementation specifics from the main experiments in \S\ref{sec:5}.
For SC, we generate 3 samples for each question since our method is also set to 3 paths. For our method, we recall top-3 statements in AAE recall and generate one CoT for each enhanced prompt.

\section{Additional Experiments on the Main Experiment}
\label{append:main_result}
We also repeat the experiments on Gemma2-9B and report the results in Table \ref{tab:main_append}, which demonstrates the effectiveness of our method.

\begin{table*}[htbp]
    \centering
    \begin{tabular}{lcccccc} 
            \toprule
            C. $\rightarrow$ A.  & GSM  & AQuA & PW & PQA & WINO & SIQA \\
            \midrule
             \ding{51} $\rightarrow$ \ding{51}& 11 & 3 & 13 & 22 & 16 & 31 \\
             \ding{51} $\rightarrow$ \ding{55} & 0 &  0  & 5 & 1 & 7 & 0\\
             \ding{55} $\rightarrow$ \ding{51} & 0 & 7 & 23 & 19 & 6 & 0 \\
             \ding{55} $\rightarrow$ \ding{55} & 39 &  40 & 9 & 8 & 21 & 19  \\
            \bottomrule
    \end{tabular}
        \caption{Inconsistency statistics between CoTs and answers on Llama2-13B.}\label{tab:llama2_faith}
\end{table*}

\begin{table*}[htbp]
\centering
\begin{tabular}{lcccccc}
\toprule
\multirow{2}{*}{\textbf{Method}} & \multicolumn{3}{c}{\textbf{ProofWriter}} & \multicolumn{3}{c}{\textbf{ProntoQA}} \\
\cmidrule(r){2-4} \cmidrule(r){5-7}
& Acc & BS & FBS & Acc & BS & FBS   \\
\midrule
 CoT & 65.0  & 56.6 & 52.9 & 77.0 & 62.7 & 57.7 \\ 
 SC & 31.0  & 54.0 & 50.3 & 81.0 & 64.5 & 60.5 \\ 
% SC@10 & 69.6  & 42.4 & 47.0 & 96.0 & 61.2 & 61.6 \\ 
LtM  & 55.0 & 55.4 & 51.8 & 90.0 & 71.0 & 67.6 \\
 SR  & 18.5 & 43.1 & 58.6 & 56.5 & 45.3 & 51.9\\
\midrule
\textbf{Ours} & \textbf{65.0} & \textbf{60.7} & \textbf{56.3} & \textbf{92.5} & \textbf{71.2} & \textbf{69.5}  \\
 - AAE Recall & 27.5 & 54.2 & 50.3 & 89.0 & 64.5 & 61.9\\
 - IG Vote & 58.5 & 57.8 & 52.8 & 74.5 & 65.9 & 60.6\\ 
\bottomrule

\end{tabular}
\caption{Results of our main experiment on Gemma2-9B, the best results are highlighted in \textbf{bold}.}
\label{tab:main_append}
\end{table*}

\end{document}